\documentclass[letterpaper]{article} 
\usepackage{aaai23}  
\usepackage{times}  
\usepackage{helvet}  
\usepackage{courier}  
\usepackage[hyphens]{url}  
\usepackage{graphicx} 
\usepackage{natbib}  
\usepackage{caption} 
\usepackage{amsfonts,bm}
\usepackage{times}
\usepackage{latexsym}
\usepackage{url}
\usepackage{algorithm}
\usepackage{algpseudocode}
\usepackage{amsmath}
\usepackage{amsthm}
\usepackage{multirow}
\usepackage{colortbl, xcolor}
\usepackage{tabularx,paralist}
\usepackage{tcolorbox}
\usepackage{adjustbox}
\usepackage{transparent}
\usepackage{booktabs}
\usepackage{graphicx}
\usepackage{tikz-dependency}
\usepackage{verbatim}
\usepackage{amssymb}
\usepackage{caption, subcaption}
\usepackage[T1]{fontenc}
\usepackage[utf8]{inputenc}
\usepackage{microtype}

\DeclareMathOperator*{\argmin}{arg\,min}

\newtheorem{definition}{Definition}

\newcommand{\norm}[1]{\left\|#1\right\|}

\definecolor{grey}{rgb}{0.1,0.1,0.1}

\newcommand{\rightt}{$\checkmark$}

\usepackage{newfloat}
\usepackage{listings}
\DeclareCaptionStyle{ruled}{labelfont=normalfont,labelsep=colon,strut=off} 
\floatstyle{ruled}
\newfloat{listing}{tb}{lst}{}
\floatname{listing}{Listing}
\title{SEAT: Stable and Explainable Attention}
\author {
    Lijie Hu \equalcontrib \textsuperscript{,\rm 1},
    Yixin Liu \equalcontrib \textsuperscript{,\rm 2},
    Ninghao Liu \textsuperscript{\rm 3},
    Mengdi Huai \textsuperscript{\rm 4},
    Lichao Sun \textsuperscript{\rm 2},
    Di Wang \textsuperscript{\rm 1, \rm 5, \rm 6}
}
\affiliations {
    \textsuperscript{\rm 1} King Abdullah University of Science and Technology, \textsuperscript{\rm 2} Lehigh University\\
    \textsuperscript{\rm 3} University of Georgia, \textsuperscript{\rm 4} Iowa State University, \textsuperscript{\rm 5} Computational Bioscience Research Center \\
    \textsuperscript{\rm 6} SDAIA-KAUST Center of Excellence in Data Science and Artificial Intelligence \\
    \{lijie.hu, di.wang\}@kaust.edu.sa, \{yila22, lis221\}@lehigh.edu, ninghao.liu@uga.edu, mdhuai@iastate.edu
}

\begin{document}

\maketitle

\begin{abstract}
Currently, attention mechanism becomes a standard fixture in most state-of-the-art natural language processing (NLP) models, not only due to outstanding performance it could gain, but also due to plausible innate explanation for the behaviors of neural architectures it provides, which is notoriously difficult to analyze. However, recent studies show that attention is unstable against randomness and perturbations during training or testing, such as random seeds and slight perturbation of embedding vectors, which impedes it from becoming a faithful explanation tool. Thus, a natural question is whether we can find some substitute of the current attention which is more stable and could keep the most important characteristics on explanation and prediction of attention. In this paper, to resolve the problem, we provide a first rigorous definition of such alternate namely SEAT (\underline{\textbf{S}}table and \underline{\textbf{E}}xplainable \underline{\textbf{At}}tention). Specifically, a SEAT should has the following three properties: (1) Its prediction distribution is enforced to be close to the distribution based on the vanilla attention; (2) Its top-$k$ indices have large overlaps with those of the vanilla attention; (3) It is robust w.r.t perturbations, i.e., any slight perturbation on SEAT will not change the prediction distribution too much, which implicitly indicates that it is stable to randomness and perturbations. Moreover we propose a method to get a SEAT, which could be considered as an ad hoc modification for the canonical attention. Finally, through intensive experiments on various datasets, we compare our SEAT with other baseline methods using RNN, BiLSTM and BERT architectures via six different evaluation metrics for model interpretation, stability and accuracy.  Results show that SEAT is more stable against different perturbations and randomness while also keeps the explainability of attention, which indicates it is a more faithful explanation. 
Moreover, compared with vanilla attention, there is almost no utility (accuracy) degradation for SEAT. 
\end{abstract}

\vspace{-10pt}
\section{Introduction}

\input{figures/examples/setup}

\begin{figure}[!t]
\centering
\includegraphics[width=1.0\columnwidth]{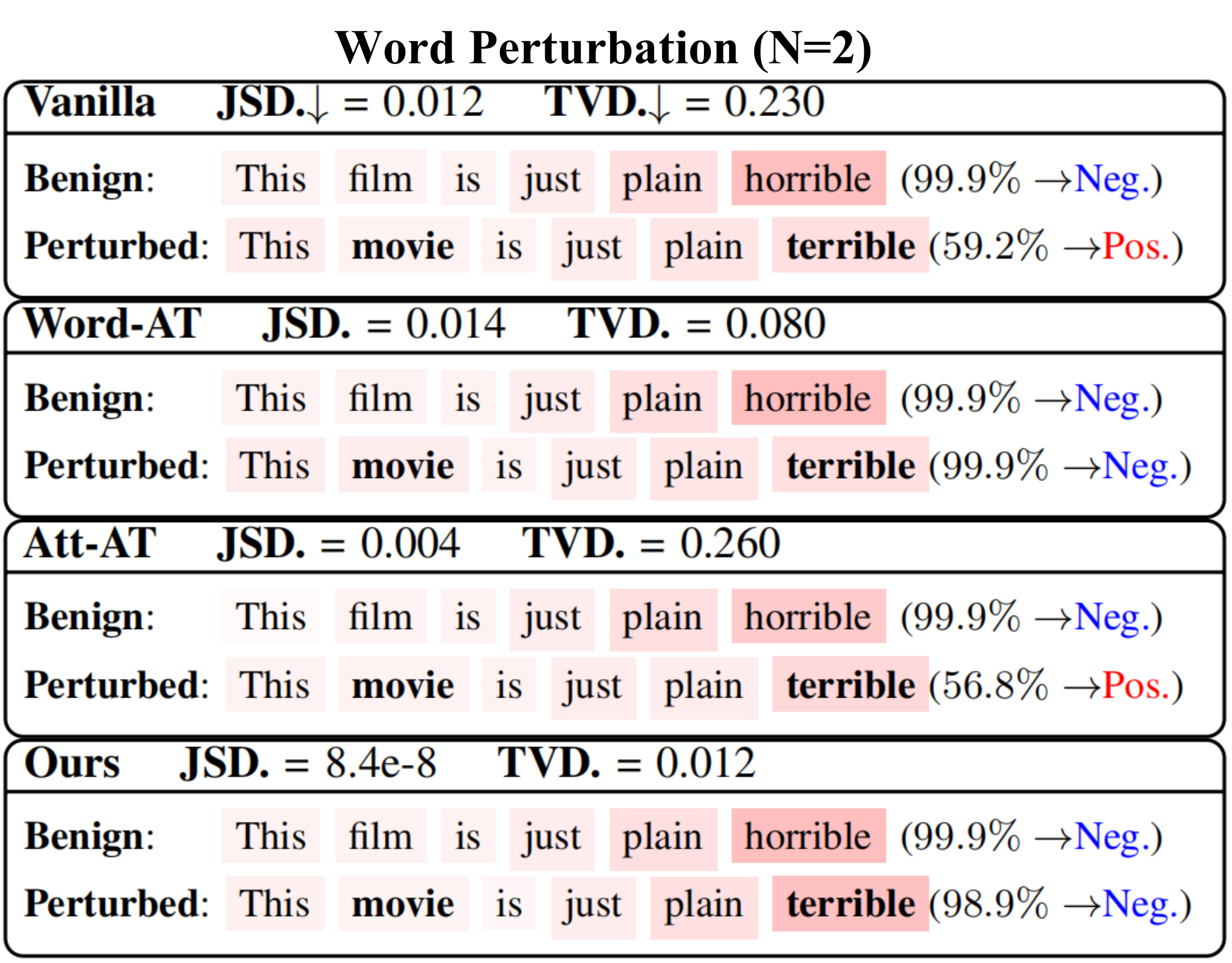}
\caption{An example demonstrates stability of prediction and attention heat map trained with different methods under word perturbation in sentiment classification task. There are four methods: Vanilla attention, Word-AT, Att-AT with and our SEAT. JSD and TVD are divergence to measure stability of explainablity and prediction distribution (see Experiments section for details). We use the closest synonyms to replace original word in sentence as word perturbation. $N$ denotes number of replaced words. We can see explanation (heat map) and prediction are changed in other methods.}
\label{example-word}
\vspace{-10pt}
\end{figure}

As deep neural networks have demonstrated great success in various natural language processing (NLP) tasks ~\cite{otter2020survey}, to further establish trust, the ability to interpret these deep models is receiving an increasing amount of attention. Recently, a number of interpretation techniques have been developed to understand the decision of deep NLP models ~\cite{ribeiro2016should, vaswani2017attention, dong2019editnts}. Among them, the \textit{attention mechanism} has been a near-ubiquitous component of modern architecture for NLP deep models. Different from post-hoc interpretation ~\cite{du2019techniques}, attention weights are often regarded as providing the "inner-workings" of models ~\cite{choi2016retain, martins2016softmax, lei2017interpretable}. For instance, each entry of the attention vector
could point us to relevant information discarded by the neural network or to irrelevant elements (tokens) of the input source that have been factored in ~\cite{galassi2020attention}.

Despite its wide adoption, attention mechanism has been questioned as being a {\bf faithful} interpretation scheme due to its instability. 
For example, ~\citet{wiegreffe2019attention} show that attention is unstable, as different model initialization could cause different attention distributions given the same input (see Fig. \ref{fig:exampleseed} in Appendix for an example). Besides the randomness during training,   attention is also fragile to various perturbations during testing. For example, in Fig. \ref{example-word}, we can see attention may give wrong predictions and its explanation heat map will be changed significantly even there is just two simple synonymous replacement perturbations. 
In addition, perturbation on embedding will also affect the prediction distribution and the explainability of attention (see details in Appendix Fig. \ref{fig:exampleembedding}). Actually, instability has been identified as a common issue of interpretation methods in deep models. Generally speaking, an unstable interpretation makes it easy to be influenced by noises in data, thus impeding users from understanding the inherent rationale behind model predictions. Moreover, instability reduces the reliability of interpretation as a diagnosis tool of models, where small carefully-crafted perturbation on input could dramatically change the interpretation result ~\cite{ghorbani2019interpretation, Dombrowski-etal19geometry, yeh2019fidelity}. Thus, stability now becomes an important factor for faithful interpretations. Based on the above facts, a natural question is can we make or modify the attention mechanism more faithful by improving its stability while also keep the most important characteristics on explanation and prediction of attention? 

In order to make the vanilla attention towards a more faithful interpretation, we first need to give a rigorous definition of such "stable attention". Intuitively, such "stable attention" should has the following three properties for any input: (1) Its prediction distribution is close to it of the vanilla attention which ensures that it keeps the outstanding performance of attention. (2) The top-$k$ indices of "stable attention" and the vanilla attention should have large overlaps, which guarantees that such "stable attention" inherits the interpretability from original attention mechanism. (3) It is stable, i.e., it is robust to any randomness and perturbations during training and testing. Based on the above three criteria, in this paper we present a formal definition of such substitute of attention namely SEAT (\underline{\textbf{S}}table and \underline{\textbf{E}}xplainable \underline{\textbf{At}}tention). Specifically, our contributions can be summarized as follows.

\begin{enumerate}
    \item We first provide a rigorous mathematical definition of SEAT. Specifically, to keep property (1), SEAT ensures the loss between its prediction distribution (vector) and the prediction distribution (vector) based on attention is sufficiently small. For property (2), we ensure the top-$k$ indices overlaps between SEAT and vanilla attention are large enough. For property (3), SEAT guarantees  some perturbations on it will not change the prediction distribution too much, which implicitly ensures  it is robust to randomness and perturbations during training and testing. 
    \item In the second part of the paper, we propose a method to find a SEAT. Specifically, we present a min-max stochastic optimization problem whose objective function involves three terms, which correspond to the above three properties. However, the main difficulty is that the term induced by
     property (2) is non-differentiable, which impedes us from using gradient descent based methods. To address this issue, we also propose a surrogate  loss function of top-$k$ overlap function as a byproduct, which can also be used in other problems. 
    \item Finally, we conduct intensive experiments on four benchmark datasets using RNN, BiLSTM and BERT to verify the above three properties of the SEAT found by our method. Particularly, using two metrics,  we first show our SEAT is more stable than other baselines via three different perturbations or randomness: random seeds, embedding vector perturbation and input token perturbation.
    We also use three recent evaluation metrics on model interpretability to measure our method. Results  reveal our SEAT is a more faithful interpretation. Besides, we also compare F1 score of our SEAT and other baselines, which shows that compared with vanilla attention there is almost no accuracy degradation for SEAT. See Figure \ref{example-word} for comparison between SEAT and other methods. 
\end{enumerate}

\vspace{-10pt}
\section{Related Work}
\paragraph{Stability and robustness in attention.}
There exists some work studying or improving either the stability or the robustness of attention from the explanation perspective. Recently, ~\citet{kitada2021attention} propose a method to improve the robustness to perturbation of embedding vector for attention. Specifically, they adopt the adversarial training during the training process. However, in their method, they do not consider the similarity and closeness between the their new attentions and the original ones, which means their robust attention loses the prediction performance and explainability of the original attention. Equivalently, while their adversarial training may could improve the robustness of attention, it can not be ensured to be explainable due to the ignorance of relationship with vanilla attention. ~\citet{sato2018interpretable} study using adversarial training to improve the robustness and interpretation of text. However, their work is applied to input embedding space, whose computational cost is high. Moreover, their method still cannot guarantee the closeness to attention on neither prediction or explanation, and their method cannot ensure the robustness against other randomness such as random seeds. ~\cite{mohankumar2020towards} explores to modify the LSTM cell with diversity driven training to enhance explainability and transparency of attention modules. However, it does not consider the robustness of attention, which makes their method far from a faithful method.

\paragraph{Stability in explanation techniques.} Besides attention, there are numerous works on studying stable interpretation. For example, \citet{yeh2019fidelity} theoretically analyze the stability of post-hoc interpretation approaches and proposes using smoothing to improve interpretation stability. ~\citet{jacovi2020towards} discuss high-level directions of designing reliable interpretation. However, these techniques are designed for post-hoc interpretation, which cannot be directly applied to attention mechanisms. Recently, \citet{yin2022sensitivity} introduce two metrics to measure the interpretability via sensitivity and stability. They also introduce methods to better test the validity of their evaluation metrics by designing an iterative gradient descent algorithm to get an counterfactual interpretation. But they do not consider how to improve faithfulness of explainable models. Thus, it is incomparable with our work. And in the experiments part we will use these evaluation metrics.

\vspace{-0.1in}
\section{Stable and Explainable Attention}
\subsection{Vanilla Attention}
  We first give a brief introduction on the attention mechanism \cite{vaswani2017attention}.  
Here we follow the notations in \cite{jain2019attention}. Let $x \in \mathbb{R}^{s \times |V|}$ denote the model input, composed of one-hot encoded words at each position. There is   an embedding matrix $E$ with dimension $d$. After passing through the embedding matrix $E$ we have a more dense token representation, which is represented as $x_e \in \mathbb{R}^{s \times d}$.

There is  an encoder (\textbf{Enc})-decoder (\textbf{Dec}) layer. For the encoder, it takes the embedded tokens in order and produces  $S$ number of $m$-dimensional hidden states after the \textbf{Enc} procedure, i.e., $\bm{h}(x)= \textbf{Enc}(x_e) \in \mathbb{R}^{s \times m}$ with $h_t(x)$ as the word representation  for the word at position $t$ in $x$. 

A similar function $\phi$ maps $\bm{h}(x)$ and a query $Q\in \mathbb{R}^m$ to scalar scores, and the vector of attention weight  is induced by $\bm{{w}(x)}=\text{softmax}\phi (\bm{h}(x),\bm{Q})$. Here we consider two common types of similarity functions: \textit{Additive} $\phi (\bm{h}(x),\bm{Q}) = \bm{v}^T \text{tanh} (\bm{W_1 h(x)}+\bm{W_2 Q})$ ~\cite{bahdanau2015neural} and \textit{Scaled Dot-Product} $ \phi (\bm{h},\bm{Q}) = \frac{\bm{h(x)Q}}{\sqrt{m}}$ ~\cite{vaswani2017attention}, where $\bm{v}$, $\bm{W_1}$, and $\bm{W_2}$ are model parameters.

Based on $\bm{{w}(x)}\in \mathbb{R}^s$, we then have prediction after the \textbf{Dec} procedure, i.e.,  $y(x, \bm{{w}}) = \sigma (\theta \cdot h_{w} ) \in \mathbb{R}^{|\mathcal{Y}|}$, where $h_{w} = \sum_{t=1}^s \bm{{w}_t(x)} \cdot h_t(x)$, $\sigma$ is an output activation function, $\theta$ is a parameter and $|\mathcal{Y}|$ denotes the label set size.

\vspace{-7pt}
\subsection{Stable and Explainable Attention}
\paragraph{Motivation.} As we mentioned previously, our goal is to find some "stable attention" that keeps the {\bf performance} and {\bf explainability} of attention, while is more {\bf robust against some randomness and perturbations during training and testing}. Before showing how to find it, we first need to think about what properties it should have. Actually, such a "stable attention" should has the following three properties:
\begin{enumerate}
    \item It keeps the advantage of outstanding performance of attention, i.e., we hope the prediction distribution (vector) based on the "stable attention" is almost the same as the distribution (vector) based on vanilla attention for any input $x$. Mathematically, we can use different loss functions or divergences  to measure such similarity or closeness. 
    \item A "stable attention" should also keep the explainability of vanilla attention. As we mentioned above, the explainability of attention could be revealed by the entries of the attention vector. More specifically, the rank of each entry in the attention vector determine the importance of its associated token. Thus, to ensure the explainablity, it is sufficiently to keep the order of  leading entries. Mathematically, here we can use the overlaps of top-$k$ indices between  "stable attention" and vanilla attention to measure their similarity on explainability, where $k$ is a parameter. 
    \item Such new attention should be stable. It is notable that compared with the robustness to adversarial attacks such as poisoning attack,  here our stability is more general, i.e., it should be robust against any randomness and perturbations during training and testing, such as random seeds in training, and perturbations on embedding vectors or input tokens during testing. Thus, unlike adversarial training, it is difficult to model the robustness to varies randomness or perturbations directly. To resolve the issue, as we mentioned, those randomness and perturbations will cause attention changes dramatically, which could be thought as some noise added to attention will change it significantly. Thus, if the "stable attention" is resilient to any perturbations, then this can indicate that such vector is robust to any randomness and perturbations implicitly. In total, mathematically we can model such robustness via the resilience against perturbations of "stable attention". 
\end{enumerate}

Based on the previous motivation, in the following we formally give the definition of "stable attention" called Stable and Explainable Attention (SEAT) and denoted as $\bm{\Tilde{w}}$. Since we need to use the overlaps of top-$k$ indices to measure the similarity on explainability with attention. We first provide its formal definition. 
\begin{definition}[Top-$k$ overlaps]
For vector $x\in \mathbb{R}^d$, we define the set of top-$k$ component $T_k(\cdot)$ as follow, 
\begin{equation*}
    T_k(x)=\{i: i\in [d] \text{ and } \{|\{x_j\geq x_i: j\in [d]\}|\leq k\} \}.
\end{equation*}
And for two vectors $x$, $x'$, the top-k overlap function $V_k(x, x')$  is defined by the overlapping ratio between the top-$k$ components of two vectors, i.e.,
    $V_k(x, x')=\frac{1}{k} | T_k(x) \cap T_k(x')|. $
\end{definition}

Note that in attention, $\bm{w}$ could be seen as a function of $x$. Thus, $\bm{\Tilde{w}}$ can also be seen as a function of $x$.
Moreover, since we only concern about replacing the attention vector, thus we will still follow the previous model except the procedure to produce the vector $\bm{\Tilde{w}}(x)$.
We define a SEAT as follows. 

\begin{definition}[\textbf{Stable and Explainable Attention}]\label{def:1}
We call a vector $\bm{\Tilde{w}}$ is a $(D_1, D_2, R, \alpha, \beta,  \gamma, V_k)$-Stable and Explainable Attention (SEAT) for  the vanilla attention  $\bm{w}$ if it satisfies for any $x$ 
\begin{itemize}
    \item (Closeness of Prediction) $D_1(y(x, \bm{\Tilde{w}}), y(x,  \bm{w})) \leq \gamma$ for some $\gamma\geq 0$, where $D_1$ is some loss, $y(x, \bm{\Tilde{w}}) = \sigma (\theta \cdot h_{\Tilde{w}} ) \in \mathbb{R}^{|\mathcal{Y}|}$  and $y(x, \bm{w})=\sigma (\theta \cdot h_{w} ) \in \mathbb{R}^{|\mathcal{Y}|}$;
    \item (Similarity of Explainability) $V_k({\bm{\Tilde{w}}(x)}, {\bm{w}(x)}) \geq \beta$ for some $1\geq \beta\geq 0$;
    \item (Stability) $D_2(y(x, \bm{\Tilde{w}}), y(x, \bm{\Tilde{w} + \delta})) \leq \alpha$ for all $\|\bm{\delta}\| \leq R$, where $D_2$ is some loss, $\|\cdot\|$ is a norm  and $R\geq 0$. 
\end{itemize}
\end{definition}
Note that in the previous definition there are several parameters. Specifically, $\gamma$ measures the closeness between the prediction distribution based on $\bm{\tilde{w}}$ and the prediction distribution based on vanilla attention. When $\gamma=0$, then $\bm{\tilde{w}}=\bm{w}$. Therefore we hope $\gamma$ to be as small as possible. The second condition ensures $\bm{\tilde{w}}$ has similar explainablity with attention. There are two parameters, $k$ and $\beta$. $k$ could be considered as a prior knowledge, i.e., we believe the top-$k$ indices of attention will play the most important role to make the prediction, or their corresponding $k$ tokens can almost determine its prediction. $\beta$ measures how much explainability does $\bm{\tilde{w}}$ inherit from vanilla attention. When $\beta=1$, then this means the top-$k$ order of the entries in $\bm{\tilde{w}}(x)$ is the same as it in vanilla attention. Thus,  $\beta$ should close to $1$. The term on stability involves two parameters $R$ and $\alpha$, which corresponds to the robust region and the level of stability respectively. Ideally, if $\bm{\tilde{w}}$ satisfies this condition with $R=\infty$ and $\alpha=0$, then $\bm{\tilde{w}}$ will be extremely stable w.r.t any randomness or perturbations. Thus, in practice we wish $R$ to be as large as possible and $\alpha$ to be sufficiently small. Thus, based on these discussions, we can see Definition \ref{def:1} is consistent with our above intuition on "stable attention" and thus it is reasonable.

\section{Finding a SEAT}
In the last section, we presented a rigorous definition of stable and explainable attention. To find such a SEAT, we propose to formulate a min-max optimization problem that involves the three conditions in Definition \ref{def:1}. Specifically, the formulated optimization problem takes the first condition (closeness of prediction) as the objective, and subjects to the other two conditions. Thus, we can get a rough optimization problem according to the definition. Specifically, we first have 
\begin{equation}
\label{eq:1}
\bm{\min_{\Tilde{w}}} \mathbb{E}_{x} D_1(y(x, \bm{\Tilde{w}}), y(x, \bm{w})).
\end{equation}
Equation (\ref{eq:1}) is the basic optimization goal, that is, we want to get a vector which has similar output prediction with vanilla attention for all input $x$. If there is no further constraint, then we can see the minimizer of (\ref{eq:1}) is just the vanilla attention $w$. 
We then consider constraints for this objective function: 
\begin{align}
\label{eq:2}
\forall x \text{ s.t.}  &\bm{\max}_{||\delta|| \leq R} D_2(y(x, \bm{\Tilde{w}}), y(x, \bm{\Tilde{w} + \delta}))\leq \alpha, \\
\label{eq:3}
&V_k({\bm{\Tilde{w}}(x)}, {\bm{w}(x)}) \geq \beta.
\end{align}
Equation (\ref{eq:2}) is the constraint of stability and Equation (\ref{eq:3}) corresponds to the condition of similarity of explainability. Combining equations (\ref{eq:1})-(\ref{eq:3}) and  using regularization to deal with constraints, 
we can get the following objective function. 
\begin{align}
    & \bm{\min_{\Tilde{w}}} \mathbb{E}_{x}[ D_1(y(x, \bm{\Tilde{w}}), y(x, \bm{w}))  + \lambda_2 (\beta - V_k({\bm{\Tilde{w}}(x)}, {\bm{w}(x))})\notag \\
    & +  \lambda_1 (\bm{\max}_{\bm{||\delta||} \leq R} D_2(y(x, \bm{\Tilde{w}}), y(x, \bm{\Tilde{w} + \delta}))-\alpha)], \label{eq:4}
\end{align}
where $\lambda_1>0$ and $\lambda_2>0$ are hyperparameters.

From now on, we convert the problem of finding a vector that satisfies the three conditions in Definition \ref{def:1} to a min-max stochastic optimization problem, where the overall objective is based on the closeness of prediction condition with constrains on stability and top-$k$ overlap.

Next we consider how to solve the above min-max optimization problem. In general, we can use the stochastic gradient descent based methods to get the solution of outer minimization, and use PSGD (Projected Stochastic Gradient Descent) to solve the inner maximization. However, the main difficulty is that the top-k overlap function $V_k(\bm{\Tilde{w}}(x), \bm{w}(x))$ is  non-differentiable, which impede us from using gradient descent. Thus, we need to consider a surrogate loss  of $-V_k(\bm{\Tilde{w}}(x), \bm{ w}(x))$. Below we provide details. 
\paragraph{Projected gradient descent to find the perturbation $\bm{\delta}$}
Motivated by \cite{madry2018towards}, we can interpret the perturbation as the attack to $\tilde{w}$ via maximizing  $\delta$. Then,  $\delta$ can be updated by the following procedure in the $k$-th iteration. 
\begin{equation*}
\bm{\delta_k} = \bm{\delta^*_{k-1}}
                 + \alpha_k \frac{1}{|B_k|}\sum_{x\in B_k} \nabla D_2(y(x, \bm{\Tilde{w}}), y(x, \bm{\Tilde{w} + \delta^*_{k-1}}));
\end{equation*}
\begin{equation*}
\bm{\delta_k^{*}} = \argmin \limits_{\bm{||\delta||} \leq R} ||\bm{\delta - \delta_k}||,
\end{equation*}
where $\alpha_k$ is a parameter of step size for PGD, $B_k$ is a batch and $|B_k|$ is the batchsize. Using this method, we can derive the optimal $\delta^{*}$ in the $t$-th iteration for the inner optimization. Specifically, we find a $\delta$ as  the maximum tolerant of perturbation w.r.t $\bm{\Tilde{w}}$ in the $t$-th iteration.
 
\begin{algorithm}[t]
    \caption{Finding a SEAT \label{alg:1}}
    \begin{algorithmic}[1]
        \State Initialize $\bm{\Tilde{w}}_0$. 
        \For {$t=1, 2, \cdots, T$}
         \State Initialize $\bm{\delta_0}$.
            \For {$k=1, 2, \cdots, K$}
                \State
                Update $\bm{\delta}$ using PGD, where $B_k$ is a batch
                \vspace{-10pt}
                \begin{multline*}
                    \bm{\delta_k} =   \bm{\delta_{k-1}}
                 \\ + \alpha_k \frac{1}{|B_k|}\sum_{x\in B_k}\nabla D_2(y(x, \bm{\Tilde{w}}_{t-1}), y(x, \bm{\Tilde{w}_{t-1} + \delta_{k-1}})).
                \end{multline*}
                \vspace{-10pt}
                \State $\bm{\delta_k^{*}} = \argmin \limits_{\bm{||\delta||} \leq R} ||\bm{\delta - \delta_k}||$.
            \EndFor
            \State Update $\bm{\Tilde{w}}$ using Stochastic Gradient Descent, where $B_t$ is a batch 
            \vspace{-10pt}
            \begin{equation*}
            \begin{split}
                &\bm{\Tilde{w}_{t}} = \bm{\Tilde{w}_{t-1}} 
                 -\frac {\eta_{t}}{|B_t|}\sum_{x\in B_t}[ \nabla D_1(y(x, \bm{\Tilde{w}_{t-1}}), y(x, \bm{w})) \\
                 &- \lambda_1 \nabla D_2(y(x, \bm{\Tilde{w}_{t-1}}),y(\bm{\Tilde{w}_{t-1}+\delta_K^{*}}))\\ &- \lambda_2 \nabla \mathcal{L}_{Topk}(\bm{w},\bm{\Tilde{w}}_{t-1})].
            \end{split}    
            \end{equation*}
            \vspace{-5pt}
        \EndFor
        \State {\bfseries Return:} $\bm{\Tilde{w}^*}=\bm{\Tilde{w}_T}$.
    \end{algorithmic}
\end{algorithm}
\vspace{-7pt}

\paragraph{Top-$k$ overlap surrogate loss}
Now we seek to design a surrogate loss $\mathcal{L}_{TopK}(\tilde{\boldsymbol{w}}, \boldsymbol{w})$ for $-V_{k}(\tilde{\boldsymbol{w}}, \boldsymbol{w})$ which can be used in training. To achieve this goal, one possible naive surrogate objective might be some distance (such as $\ell_1$-norm) between $\tilde{\boldsymbol{w}}$  and ${\boldsymbol{w}}$, e.g., $L(\tilde{\boldsymbol{w}})=||\tilde{\boldsymbol{w}}-\boldsymbol{w}||_1$.  Such surrogate objective seems like could ensure the top-$k$ overlap when we obtain the optimal or near-optimal solution (i.e., $w=\arg\min L(\tilde{\boldsymbol{w}})$ and $w\in \arg\min -V_{k}(\tilde{\boldsymbol{w}}, \boldsymbol{w})$). However, it lacks consideration of the top-$k$ information which makes it as a loose surrogate loss. Since we only need to ensure high top-$k$ indices overlaps between $\tilde{\boldsymbol{w}}$ and ${\boldsymbol{w}}$, one improved method is  minimizing the distance between  $\tilde{\boldsymbol{w}}$  and ${\boldsymbol{w}}$ constrained on the top-$k$ entries only instead of the whole vectors, i.e., $||\boldsymbol{w}_{S_w^{k}} - \boldsymbol{\tilde{w}}_{S_{w}^{k}}||_1$, where $\boldsymbol{w}_{S_w^{k}},\boldsymbol{\tilde{w}}_{S_{w}^{k}} \in \mathbb{R}^k$ is  the  vector $w$ and $\boldsymbol{\tilde{w}}$ constrained on the indices set $S_w^{k}$ respectively and $S_w^{k}$ is the top-k indices set of $w$. Since there are two top-k indices sets, one is for $\tilde{\boldsymbol{w}}$ and  the other one is for ${\boldsymbol{w}}$, here we need to use both of them to involve the top-k indices formation for both vectors. Thus, based on our above idea, our surrogate can be written as follow, 
\vspace{-10pt}
\begin{equation}
\small 
\label{eq:5}
\mathcal{L}_{Topk}(\boldsymbol{w}, \boldsymbol{\tilde{w}}) = \frac{1}{2k}  ( ||\boldsymbol{w}_{S_w^{k}} - \boldsymbol{\tilde{w}}_{S_{w}^{k}}||_1 + || \boldsymbol{\tilde{w}}_{S_{\tilde{w}}^{k}} - \boldsymbol{w}_{S_{\tilde{w}}^{k}} ||_1).  
\end{equation}
Note that besides the $\ell_1$-norm, we can use other norms. However, in practice we find $\ell_1$-norm achieves the best performance. Thus, throughout the paper we only use $\ell_1$-norm. 

\begin{table*}[t]
\renewcommand{\arraystretch}{0.8}
\centering
\resizebox{1.0\linewidth}{!}{

\begin{tabular}{cccccccccccccc}
\toprule
\multirow{2}{*}{\textbf{Model}}  & \multirow{2}{*}{\textbf{Method}} & \multicolumn{3}{c}{\textbf{Emotion}}                      & \multicolumn{3}{c}{\textbf{SST}}   & \multicolumn{3}{c}{\textbf{Hate}}  & \multicolumn{3}{c}{\textbf{RottenT}}  \\ 
\cmidrule(lr){3-5}\cmidrule(lr){6-8}\cmidrule(lr){9-11}\cmidrule(lr){12-14}
                        &                         & \textbf{JSD$\downarrow$} & \textbf{TVD$\downarrow$} & \textbf{F1$\uparrow$} & \textbf{JSD}  & \textbf{TVD}  & \textbf{F1}  & \textbf{JSD}  & \textbf{TVD}  & \textbf{F1} & \textbf{JSD}  & \textbf{TVD}  & \textbf{F1}   \\ 
\midrule
\multirow{7}{*}{\textbf{RNN}}    & Vanilla                 & 0.002           & 20.145          & 0.663        & 0.019    & 19.566 & 0.811 & 0.009    & 15.576 & 0.553 & 0.008    & 19.139 & 0.763    \\
                        & Word-AT                 & 0.028           & 1.824           & 0.627        & 0.016    & 1.130  & 0.798 & 0.026    & 1.170  & 0.527 & 0.037    & 1.381  & 0.741    \\
                        & Word-iAT                & 0.042           & 2.691           & 0.653        & 0.023    & 1.277  & \textbf{0.815} & 0.022    & 1.049  & 0.523 & 0.054    & 1.336  & 0.766    \\
                        & Attention-RP            & 0.025           & 3.276           & 0.671        & 0.028    & 2.042  & 0.792 & 0.025    & 2.672  & 0.554 & 0.009    & 3.691  & \textbf{0.770}    \\
                        & Attention-AT            & 0.055           & 2.716           & 0.665        & 0.047    & 2.394  & 0.782 & 0.031    & 2.210  & 0.528 & 0.068    & 4.234  & 0.755    \\
                        & Attention-iAT           & 0.017           & 3.654           & 0.645        & 0.048    & 2.653  & 0.746 & 0.039    & 2.264  & 0.533 & 0.054    & 1.594  & 0.753    \\
                        \rowcolor{grey!20} & SEAT(\textbf{Ours})     & \textbf{3.81E-08}        & \textbf{1.750}           & \textbf{0.672}        & \textbf{2.75E-07} & \textbf{1.099}  & 0.813 & \textbf{1.79E-09} & \textbf{0.908}  & \textbf{0.579} & \textbf{6.46E-07} & \textbf{1.178}  & 0.763    \\ 
\midrule
\midrule
\multirow{7}{*}{\textbf{BiLSTM}} & Vanilla                 & 0.002           & 23.447          & 0.612        & 0.027    & 18.640 & \textbf{0.809} & 0.060    & 15.633 & 0.524 & 0.009    & 20.125 & 0.764    \\
                        & Word-AT                 & 0.050           & 1.927           & 0.662        & 0.020    & 0.810  & 0.798 & 0.084    & 1.537  & 0.538 & 0.031    & 1.071  & 0.757    \\
                        & Word-iAT                & 0.058           & 1.139           & 0.640        & 0.034    & 1.037  & 0.802 & 0.091    & 1.590  & 0.530 & 0.045    & 1.218  & 0.765    \\
                        & Attention-RP            & 0.031           & 1.326           & 0.642        & 0.034    & 1.267  & 0.772 & 0.052    & 1.299  & 0.522 & 0.066    & 1.412  & 0.764    \\
                        & Attention-AT            & 0.076           & 1.541           & \textbf{0.672}        & 0.028    & 1.661  & 0.779 & 0.057    & 1.504  & 0.523 & 0.079    & 2.044  & 0.766    \\
                        & Attention-iAT           & 0.033           & 1.267           & 0.651        & 0.034    & 1.528  & 0.801 & 0.062    & 2.256  & 0.525 & 0.076    & 1.751  & \textbf{0.777}    \\
                        \rowcolor{grey!20} &  SEAT(\textbf{Ours})     & \textbf{1.23E-08}  & \textbf{0.736}           & 0.670        & \textbf{1.80E-08} & \textbf{0.777}  & 0.802 & \textbf{8.49E-09} & \textbf{1.030}  & \textbf{0.543} & \textbf{2.57E-08} & \textbf{0.885}  & 0.771    \\ 
\midrule
\midrule
\multirow{7}{*}{\textbf{BERT}}   & Vanilla                 & 0.024           & 2.127           & \textbf{0.721}        & 0.005    & 2.605  & 0.912 & 0.036    & 1.771  & 0.493 & 0.010    & 2.500  & \textbf{0.845}    \\
                        & Word-AT                 & 0.085           & 0.060           & 0.694        & 0.267    & 0.055  & 0.900 & 0.170    & 0.043  & \textbf{0.554} & 0.510    & 0.036  & 0.826    \\
                        & Word-iAT                & 0.584           & 0.029           & 0.694        & 0.241    & 0.054  & 0.895 & 0.166    & 0.049  & 0.496 & 0.480    & 0.049  & 0.844    \\
                        & Attention-RP            & 0.035           & 0.232           & 0.657        & 0.086    & 0.127  & 0.893 & 0.079    & 0.277  & \textbf{0.554} & 0.078    & 0.142  & 0.817    \\
                        & Attention-AT            & 0.067           & 0.119           & 0.707        & 0.005    & 0.156  & 0.907 & 0.031    & 0.230  & 0.510 & 0.041    & 0.189  & 0.818    \\
                        & Attention-iAT           & 0.096           & 0.222           & 0.684        & 0.129    & 0.200  & \textbf{0.915} & 0.074    & 0.271  & 0.512 & 0.108    & 0.183  & 0.831    \\
                        \rowcolor{grey!20} &  SEAT(\textbf{Ours})     & \textbf{4.70E-07}        & \textbf{0.002}           & 0.713        & \textbf{2.77E-07} & \textbf{0.036}  & 0.907 & \textbf{2.42E-06} & \textbf{0.042}  & 0.545 & \textbf{1.68E-08} & \textbf{0.003}  & 0.841   \\
\bottomrule
\end{tabular}
}
\caption{Results of evaluating embedding perturbation stability of (modified) attentions given by different methods using RNN, BiLSTM, and BERT under three metrics (JSD is for attention weight distribution, TVD is for output distribution, and F1 score is for model performance). The perturbation radius is set as $\delta_x$=1e-3. $\uparrow$ means a higher value under this metric indicates better results, and $\downarrow$ means the opposite. The best performance is \textbf{bolded}. Same symbols are used in the following tables by default. \label{golden}}
\end{table*}

\paragraph{Final objective function and algorithm} Based on the above discussion, we can derive the following overall objective function
\begin{align}
    & \bm{\min_{\Tilde{w}}} \mathbb{E}_{x} [D_1(y(x, \bm{\Tilde{w}}), y(x, \bm{w})) + \lambda_2 \mathcal{L}_{Topk}({\bm{w}(x)}, {\bm{\Tilde{w}}(x)}) \notag \\
    & +  \lambda_1 \bm{\max}_{\bm{||\delta||} \leq R} D_2(y(x, \bm{\Tilde{w}}), y(x, \bm{\Tilde{w} + \delta}))], \label{eq:6}
\end{align}
where $\mathcal{L}_{Topk}(\bm{w}, \bm{\Tilde{w}})$ is defined in (\ref{eq:5}). Based on the previous idea, we propose Algorithm \ref{alg:1} to solve (\ref{eq:6}). 

\begin{figure*}[ht]
    \centering
    \small
    \begin{tabular}{cccc}
        \includegraphics[width=3.5cm,height=0.12\textheight]{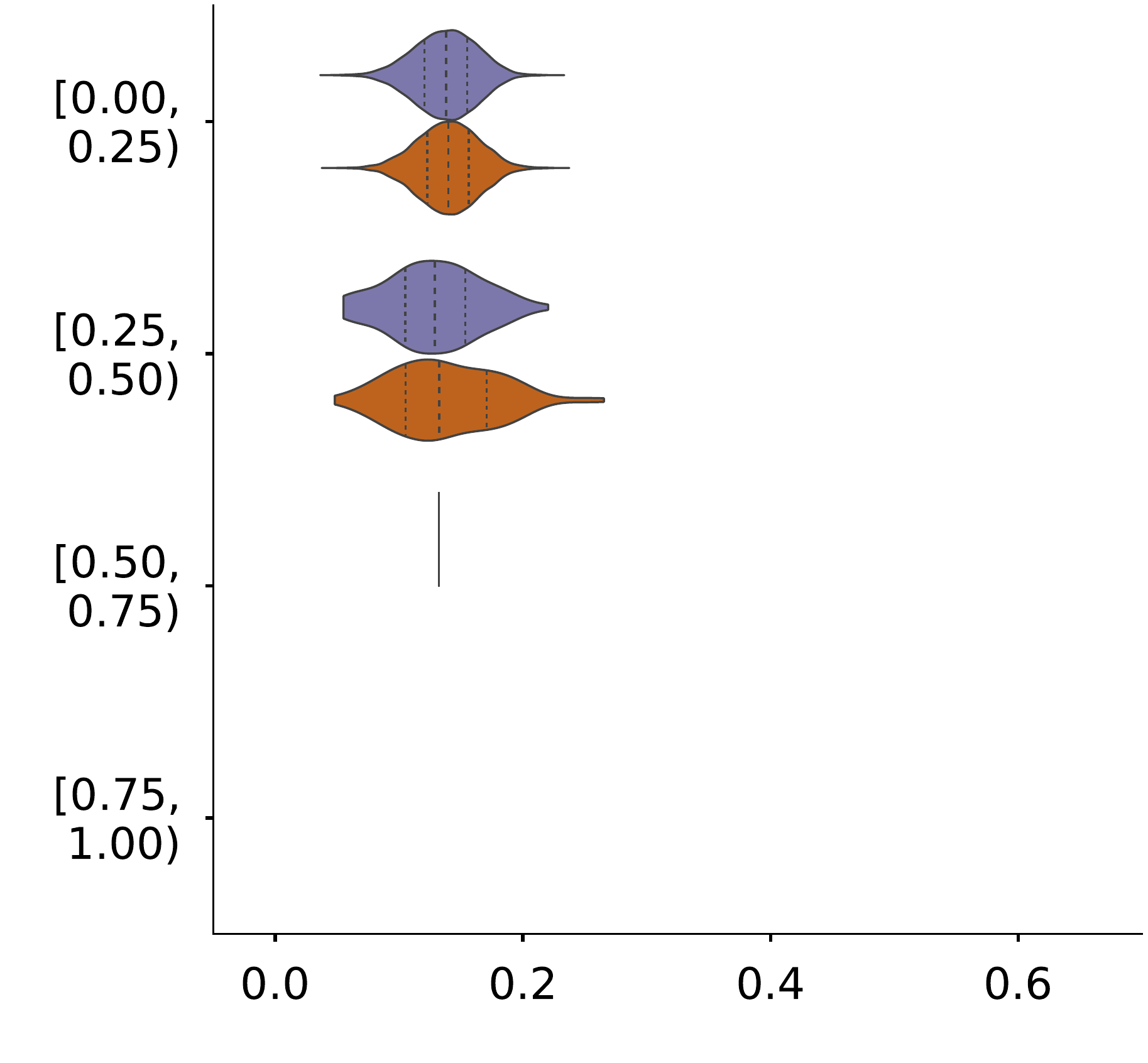}
        & 
        \includegraphics[width=3.5cm,height=0.12\textheight]{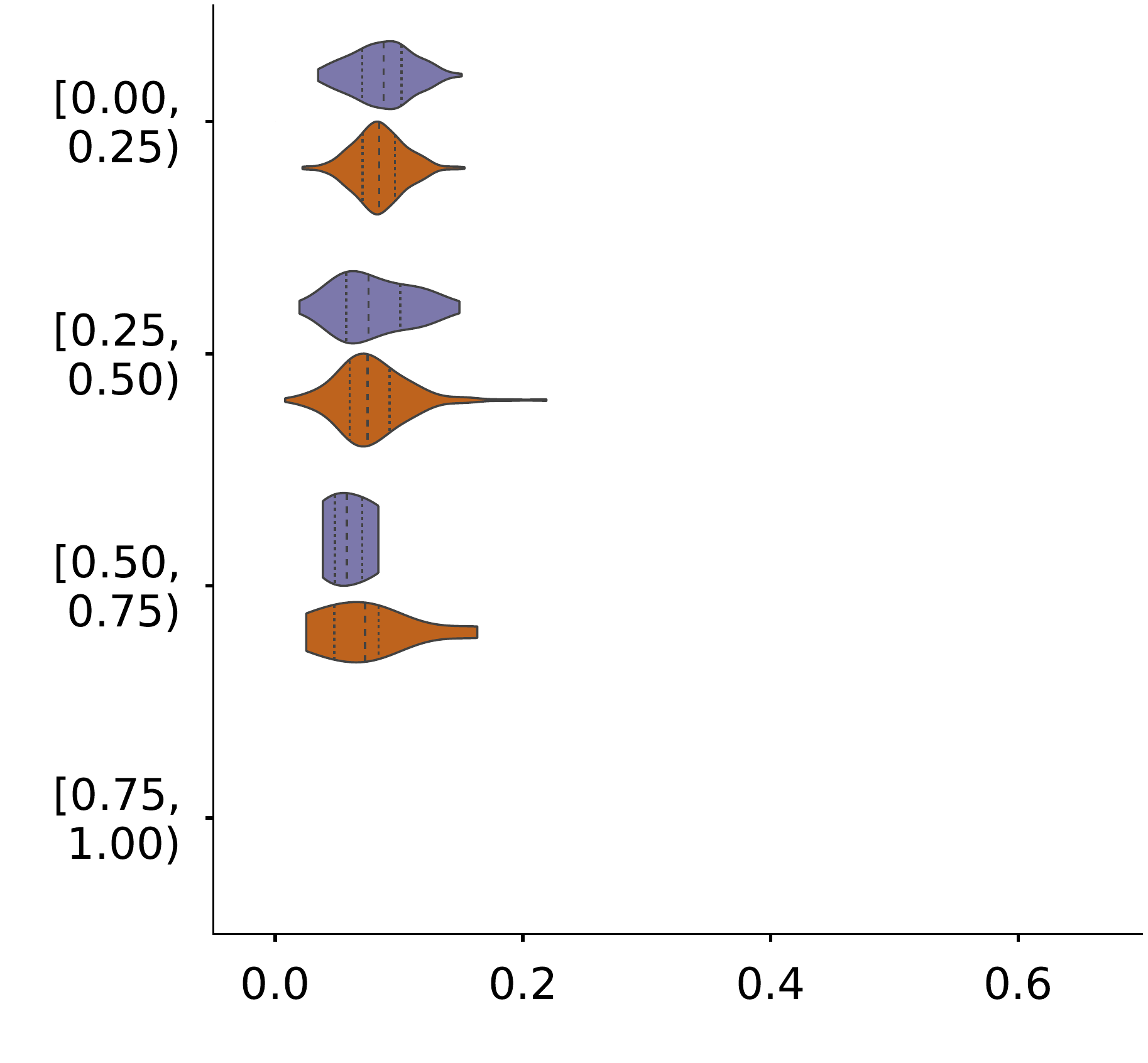} &
        \includegraphics[width=3.5cm,height=0.12\textheight]{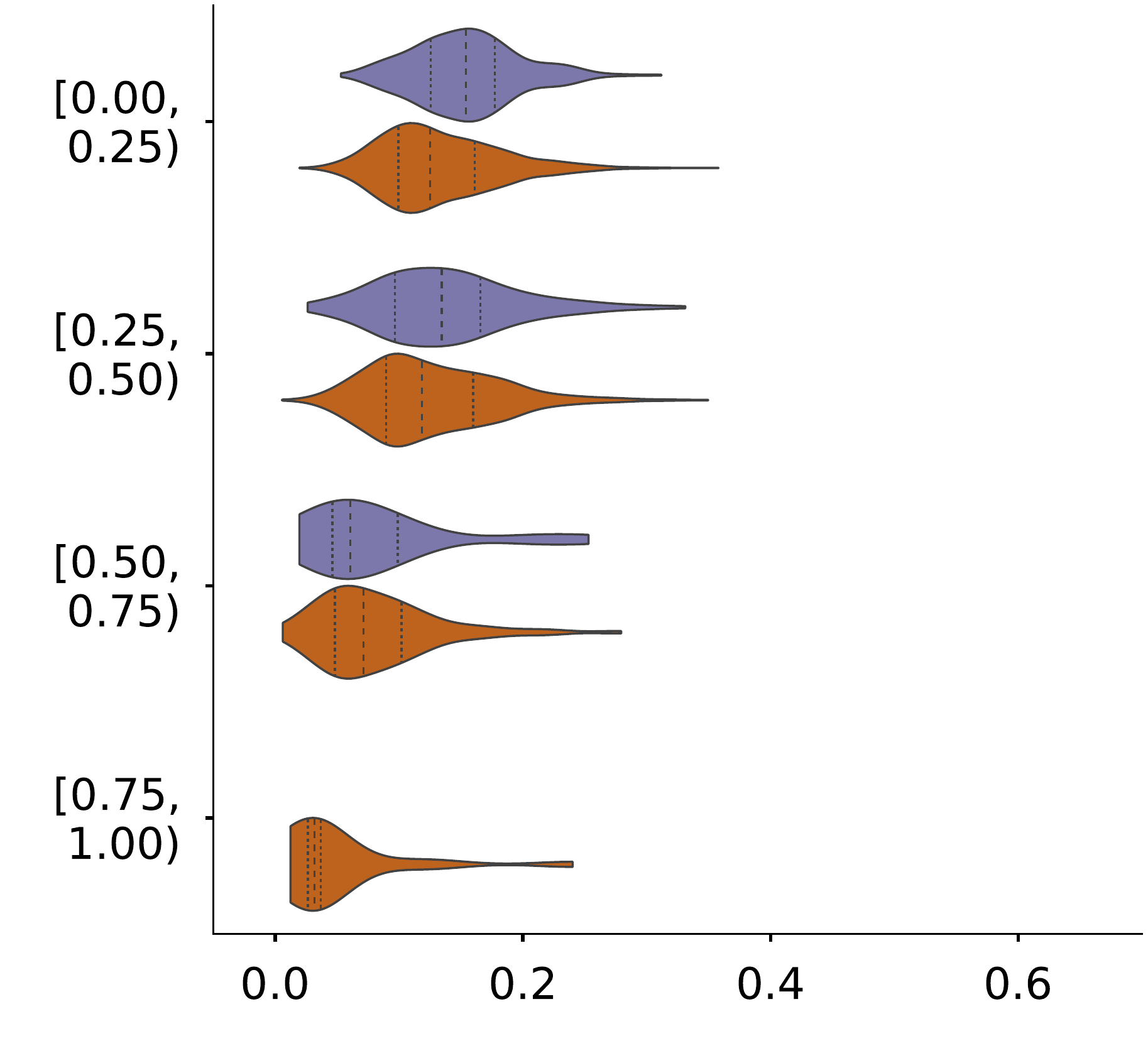}
        & 
        \includegraphics[width=3.5cm,height=0.12\textheight]{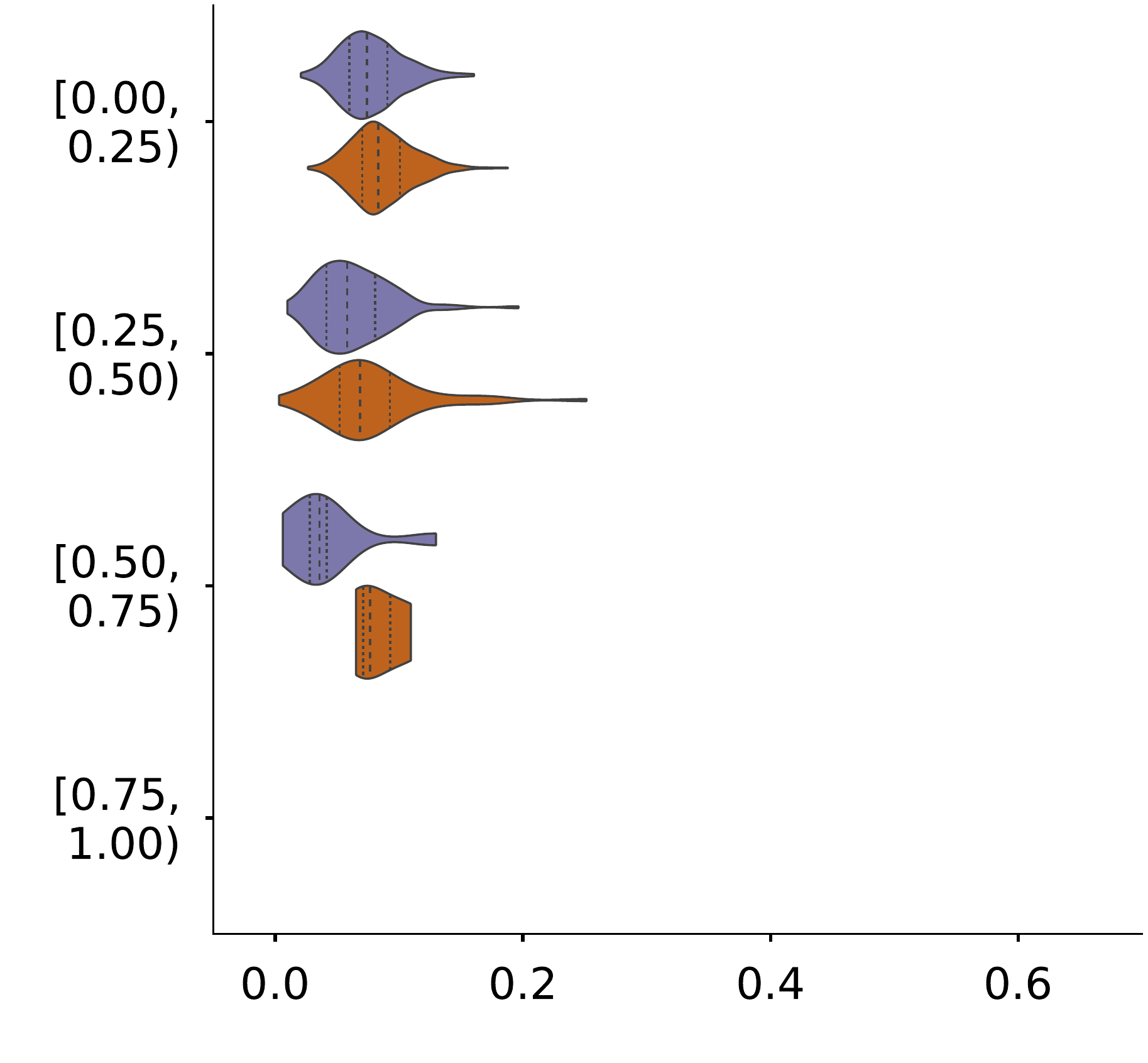} \\
        (a) Baseline-\textsc{Emotion} & (b) Baseline-SST  &
        (c) Baseline-Hate & (d) Baseline-RottenT \\
        \includegraphics[width=3.5cm,height=0.12\textheight]{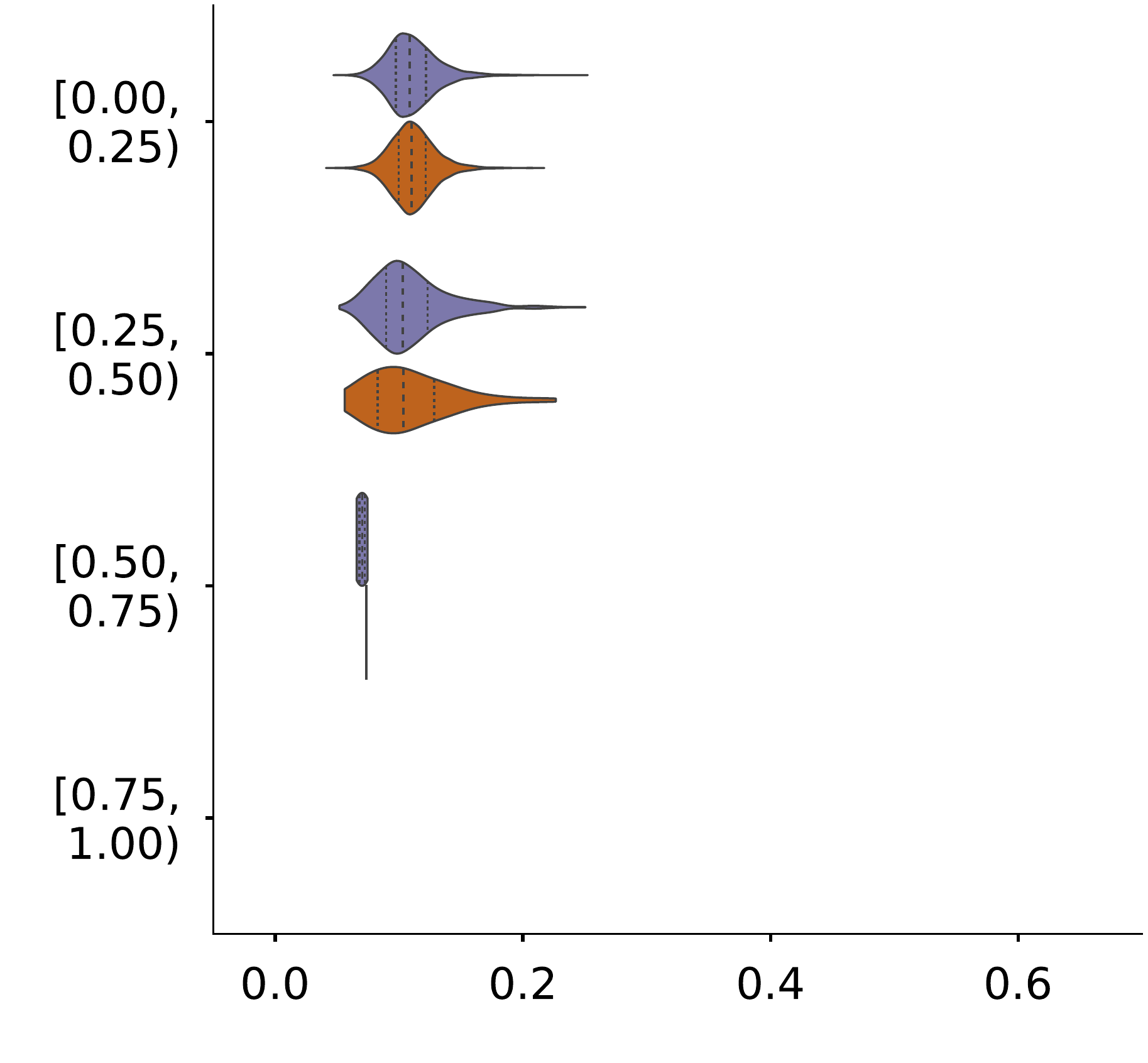}
        & 
        \includegraphics[width=3.5cm,height=0.12\textheight]{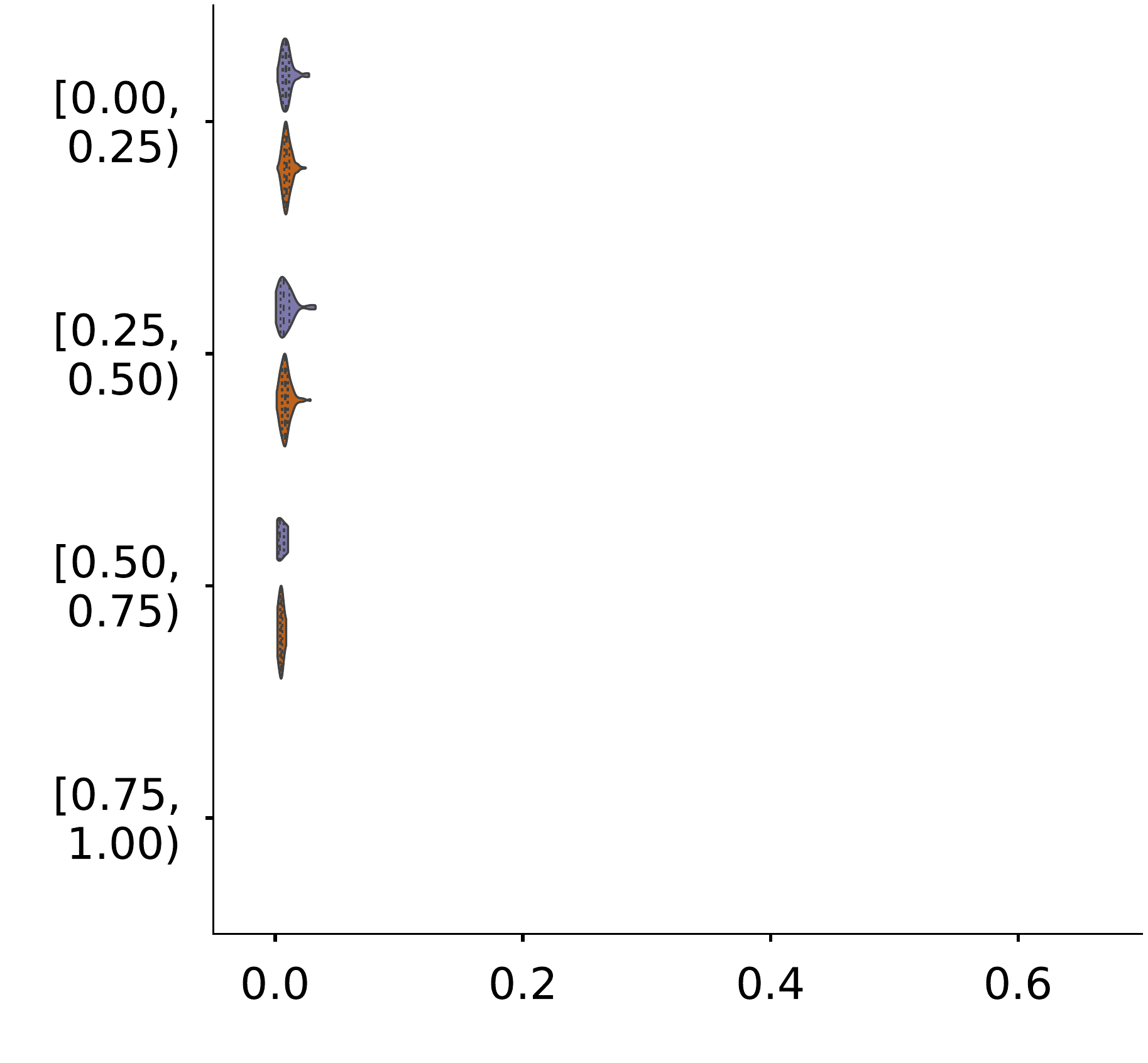} &
        \includegraphics[width=3.5cm,height=0.12\textheight]{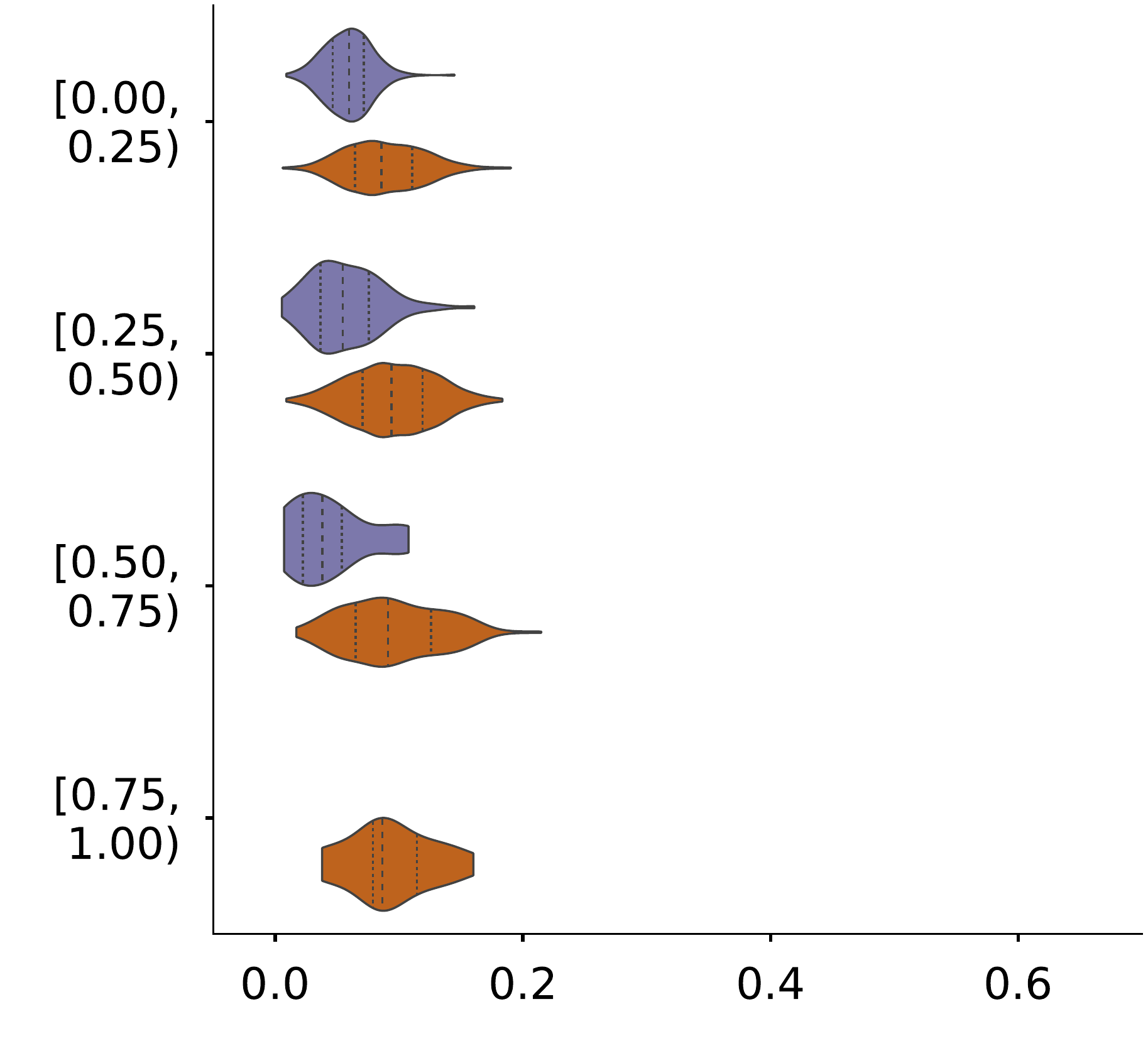}
        & 
        \includegraphics[width=3.5cm,height=0.12\textheight]{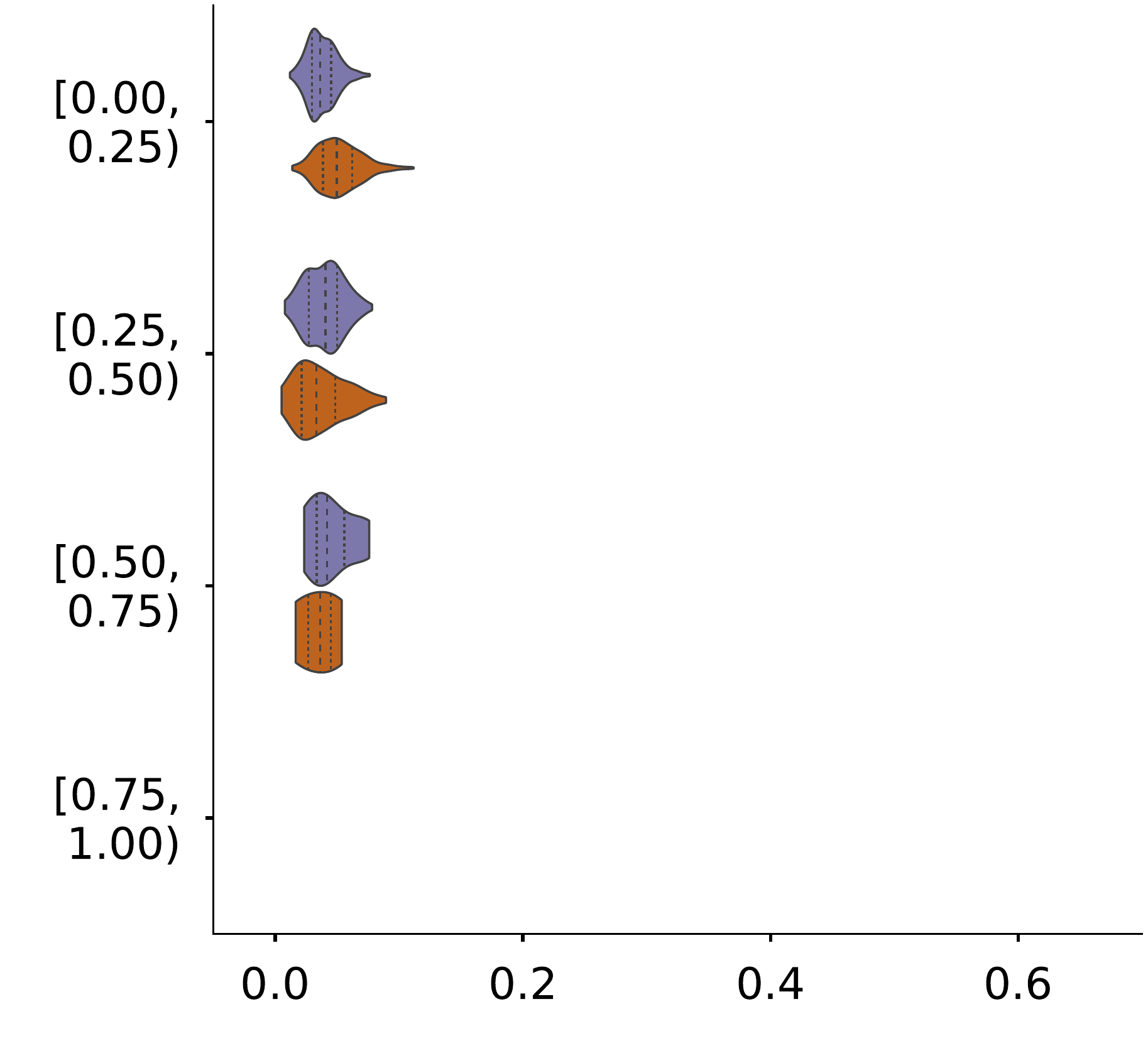} \\
        (e) Ours-\textsc{Emotion} & (f) Ours-SST  &
        (g) Ours-Hate & (h) Ours-RottenT \\
    \end{tabular}
    \caption{Comparison of stability against random seeds for vanilla attention and SEAT. Densities of maximum JS divergences (x-axis) as a function of the max attention (y-axis) in each instance
between its base model and models initialized on different random seeds. In each max-attention
bin, top (blue) is a negative-label instance, and bottom (red) is a positive-label instance.
\label{fig:seeds_main}}
\end{figure*}

\vspace{-10pt}
\begin{table*}[!ht]
\renewcommand{\arraystretch}{0.8}
\centering
\resizebox{1.0\linewidth}{!}{
\begin{tabular}{cccccccccccccc} 
\toprule
\multirow{2}{*}{\textbf{Model}}  & \multirow{2}{*}{\textbf{Method}} & \multicolumn{3}{c}{\textbf{IMDB}}                                                                         & \multicolumn{3}{c}{\textbf{SST}}     & \multicolumn{3}{c}{\textbf{Hate}}    & \multicolumn{3}{c}{\textbf{RottenT}}  \\ 
\cmidrule(lr){3-5}\cmidrule(lr){6-8}\cmidrule(lr){9-11}\cmidrule(r){12-14}
                        &    & \textbf{Comp.$\uparrow$} & \textbf{Suff.$\downarrow$} & \textbf{Sens.$\downarrow$} & \textbf{Comp.}    & \textbf{Suff.}    & \textbf{Sens.} & \textbf{Comp.}    & \textbf{Suff.}    & \textbf{Sens.} & \textbf{Comp.}    & \textbf{Suff.} & \textbf{Sens.}     \\ 
\midrule
\multirow{7}{*}{\textbf{RNN}}    & Vanilla                 & 0.004                          & 0.007                          & 0.131                          & 5.744    & 7.02E-04 & 0.090 & 0.009    & 0.066    & 0.138 & 4.483    & 0.026 & 0.090     \\
                        & Word-AT                 & 2.899                          & 0.018                          & 0.121                          & 5.280    & \textbf{0.000} & 0.081 & 2.408    & 0.058    & 0.137 & 2.512    & 0.031 & 0.093     \\
                        & Word-iAT                & 2.060                          & 0.010                          & 0.122                          & 5.452    & 9.35E-06 & 0.121 & 6.069    & 0.075    & 0.136 & 4.534    & 0.058 & 0.087     \\
                        & Attention-RP            & 0.099                          & 0.073                          & 0.130                          & 6.001    & 2.87E-04 & 0.085 & 3.585    & 0.080    & 0.133 & 4.637    & 0.052 & 0.094     \\
                        & Attention-AT            & 0.026                          & 0.052                          & 0.131                          & 3.118    & \textbf{0.000} & 0.088 & 2.493    & 0.372    & 0.134 & 4.713    & 0.049 & 0.096     \\
                        & Attention-iAT           & 1.994                          & 6.74E-04                       & 0.117                          & 4.788    & \textbf{0.000} & 0.091 & 5.351    & 0.126    & 0.133 & 2.435    & 0.039 & 0.086     \\
                        \rowcolor{grey!20} &  SEAT(\textbf{Ours})     & \textbf{3.281}                          & \textbf{1.04E-05}                       & \textbf{0.106}                          & \textbf{6.016}    & \textbf{0.000} & \textbf{0.076} & \textbf{6.558}    & \textbf{2.75E-04} & \textbf{0.129} & \textbf{4.796}    & \textbf{0.025} & \textbf{0.084}     \\ 
\midrule
\midrule
\multirow{7}{*}{\textbf{BiLSTM}} & Vanilla                 & 0.474                          & 0.002                          & 0.129                          & 5.182    & 0.255    & 0.086 & 4.203    & 0.112    & 0.142 & 2.966    & 0.088 & 0.092     \\
                        & Word-AT                 & 1.449                          & 0.015                          & 0.121                          & 5.167    & 8.44E-04 & 0.096 & 5.438    & 0.207    & 0.153 & 3.388    & 0.062 & 0.080     \\
                        & Word-iAT                & 0.619                          & 0.005                          & 0.127                          & 5.259    & 3.81E-05 & 0.087 & 4.568    & 0.320    & 0.145 & 3.339    & 0.078 & 0.082     \\
                        & Attention-RP            & 0.561                          & 0.002                          & 0.127                          & 2.865    & 0.007    & 0.101 & 2.248    & 0.199    & 0.148 & 4.073    & 0.200 & 0.082     \\
                        & Attention-AT            & 1.294                          & 0.051                          & 0.111                          & 2.129    & 0.004    & 0.098 & 3.220    & 0.065    & \textbf{0.140} & 4.925    & 0.216 & 0.082     \\
                        & Attention-iAT           & 0.555                          & 0.002                          & 0.127                          & 5.176    & 0.004    & 0.083 & 4.092    & 0.290    & 0.141 & 2.431    & 0.377 & 0.083     \\
                        \rowcolor{grey!20} &  SEAT(\textbf{Ours})     & \textbf{1.502}                          & \textbf{6.41E-04}                       & \textbf{0.098}                          & \textbf{5.435}    & \textbf{4.37E-06} & \textbf{0.076} & \textbf{6.240}    & \textbf{0.025}    & \textbf{0.140} & \textbf{4.941}    & \textbf{0.046} & \textbf{0.077}     \\ 
\midrule
\midrule
\multirow{7}{*}{\textbf{BERT}}   & Vanilla                 & 5.07E-04                       & 0.008                          & 0.013                          & 0.003    & 0.310    & 0.009 & 3.20E-04 & 0.401    & 0.016 & 0.001    & 0.092 & 0.010     \\
                        & Word-AT                 & 5.01E-05                       & 0.005                          & 0.016                          & 1.26E-05 & 4.47E-04 & 0.005 & 4.01E-04 & 0.014    & 0.017 & 0.005    & 0.043 & 0.016     \\
                        & Word-iAT                & 5.47E-04                       & 0.007                          & 0.017                          & 1.51E-05 & 5.67E-04 & 0.004 & 0.003    & 0.045    & 0.017 & 2.93E-04 & 0.010 & 0.009     \\
                        & Attention-RP            & 0.085                          & 0.086                          & 0.014                          & 0.002    & 0.010    & 0.011 & 0.035    & 0.034    & 0.016 & 0.017    & 0.003 & 0.010     \\
                        & Attention-AT            & 4.65E-05                       & 0.338                          & 0.016                          & 4.50E-05 & 0.441    & 0.006 & \textbf{0.004}    & 0.007    & 0.016 & 6.26E-04 & 0.032 & 0.011     \\
                        & Attention-iAT           & 0.002                          & 0.164                          & 0.015                          & 1.19E-04 & 9.07E-04 & 0.007 & 0.001    & 0.151    & 0.017 & 0.003    & 0.025 & 0.010     \\
                        \rowcolor{grey!20} & SEAT(\textbf{Ours})     & \textbf{0.160}                          & \textbf{0.002}                          & \textbf{0.012}                          & \textbf{0.497}    & \textbf{0.000} & \textbf{0.004} & \textbf{0.153}    & \textbf{0.006}    & \textbf{0.015} & \textbf{0.040}    & \textbf{0.002} & \textbf{0.008}     \\
\bottomrule
\end{tabular}
}
\caption{Results on evaluating the interpretability of different methods. We conduct experiments with three architectures, including RNN, BiLSTM, and BERT, under three criteria (comprehensiveness, sufficiency, and sensitivity score) on text classification datasets. The best performance is \textbf{bolded}.}
\label{golden-explanation}
\vspace{-12pt}
\end{table*}

\begin{table*}
\renewcommand{\arraystretch}{0.8}
\centering
\resizebox{1.0\linewidth}{!}{
\begin{tabular}{cccccccccccccc} 
\toprule
 \multirow{2}{*}{\textbf{Method}} & \multicolumn{3}{c}{\textbf{Emotion}}                                                             & \multicolumn{3}{c}{\textbf{SST}}  & \multicolumn{3}{c}{\textbf{Hate}} & \multicolumn{3}{c}{\textbf{RottenT}}  \\ 
\cmidrule(lr){2-4}\cmidrule(r){5-7}\cmidrule(lr){8-10}\cmidrule(lr){11-13}
                                  &      \textbf{JSD$\downarrow$} & \textbf{TVD$\downarrow$} & \textbf{F1$\uparrow$} & \textbf{JSD} & \textbf{TVD}   & \textbf{F1}    & \textbf{JSD}      & \textbf{TVD}   & \textbf{F1}    & \textbf{JSD}      & \textbf{TVD}   & \textbf{F1}        \\ 
\midrule
  \textbf{Vanilla}           & 0.628                        & 2.847                        & \textbf{0.721}                     & 0.315    & 3.655 & 0.912 & 0.491    & 2.004 & 0.493 & 0.585    & 3.464 & 0.845     \\
                                   \textbf{Word-AT}                 & 0.004                        & 0.022                        & 0.694                     & 0.175    & 0.065 & 0.910 & 0.111    & 0.058 & 0.546 & 0.473    & 0.056 & 0.836     \\
                                   \textbf{Word-iAT}                & 0.456                        & 0.059                        & 0.658                     & 0.213    & 0.046 & 0.912 & 0.331    & 0.044 & 0.501 & 0.488    & 0.048 & \textbf{0.852}     \\
                                   \textbf{Attention-RP}            & 0.039                        & 0.235                        & 0.657                     & 0.089    & 0.128 & 0.893 & 0.085    & 0.278 & 0.554 & 0.078    & 0.143 & 0.817     \\
                                   \textbf{Attention-AT}            & 0.082                        & 0.003                        & 0.707                     & 0.006    & 0.157 & 0.907 & 0.035    & 0.230 & 0.510 & 0.049    & 0.193 & 0.818     \\
                                   \textbf{Attention-iAT}           & 0.126                        & 0.228                        & 0.684                     & 0.147    & 0.204 & \textbf{0.915} & 0.081    & 0.271 & 0.512 & 0.136    & 0.187 & 0.831     \\
                                 \rowcolor{grey!20}  \textbf{SEAT(Ours)}    & \textbf{1.72E-09}                     & \textbf{0.001}                        & 0.716                     & \textbf{1.55E-06} & \textbf{0.028} & 0.907 & \textbf{8.69E-06} & \textbf{0.037} & \textbf{0.555} & \textbf{1.10E-05} & \textbf{0.035} & 0.847     \\ 
\bottomrule
\end{tabular}}
\caption{Results on stability and utility of attention model trained with different methods under word perturbation. BERT is used as our architecture, and the perturbation word size is set as $N=1$. JSD, TVD, and F1 are reported.}
\label{tab_main_wordp}
\vspace{-12pt}
\end{table*}

\vspace{-0.1in}
\section{Experiments}
In our experiments, we conduct extensive experiments to show the performance of our SEAT compared to six baseline methods. In the following we provide a brief introduction on experimental setup. {\bf More details about the setting, implementation and additional experimental results can be found in Appendix.} 

\paragraph{\textbf{Setup.}} First, we demonstrate the stability of SEAT under three different randomness and perturbations: (1) random seeds during training; (2) embedding perturbation during testing, and (3) word perturbation during testing.
For each method, we use Jensen-Shannon Divergence (JSD) between its attention with on perturbation and its attention under perturbation to evaluate the stability of explainablity of the learned attention. And the Total
Variation Distance (TVD) between the prediction distribution with no perturbation and prediction under perturbation is used to measure prediction stability. 

Next, in order to show explainability of SEAT, we use the recent evaluation metrics of model interpretation proposed  by~\cite{yin2022sensitivity,deyoung2020eraser}. Specifically, we use three explainablity evaluation metrics: \textit{comprehensiveness, sufficiency and sensitivity}. 

Thirdly, we compare the \textit{F1 score} of our SEAT with other baselines to verity the property of closeness of prediction. Finally, we conduct ablation study to verify the efficiency of our modules (regularizers) in objective function (\ref{eq:6}) corresponding to each condition. In Fig. \ref{fig:topk} of Appendix we also test the validity of surrogate loss for top-$k$ overlap function by comparing the  performance of our loss (\ref{eq:5}) with the true top-$k$ indices overlaps.
\vspace{-0.1in}
\paragraph{Model, Dataset and Baseline.} Following ~\cite{jain2019attention} and ~\cite{wiegreffe2019attention}, we mainly study the encoder-decoder architectures for binary classification tasks in this paper. For encoder, we consider three kinds of networks as feature extractors: RNN, BiLSTM, and BERT. For decoder, we apply one simple MLP followed by a tanh-attention layer ~\cite{bahdanau2015neural} and a softmax layer ~\cite{vaswani2017attention}. In all experiments, we use four datasets: Stanford Sentiment Treebank (SST)~\cite{socher2013recursive}, Emotion Recognition (Emotion)~\cite{emotion}, Hate~\cite{basile-etal-2019-semeval} and Rotten Tomatoes (RottenT)~\cite{pang2005seeing}. And we select the Binary Cross Entropy loss as $D_1$ and $D_2$ in (\ref{eq:6}). We compare our method with Vanilla attention~\cite{wiegreffe2019attention}, Word AT~\cite{miyato2016adversarial}, Word iAT~\cite{sato2018interpretable}, Attention RP (attention weight is trained with random perturbation), Attention AT and Attention iAT ~\cite{kitada2021attention}.

\subsection{Stability Evaluation}
\paragraph{Random seeds} 
Here we compare the stability against random seed for vanilla attention and our SEAT. Specifically,  we conduct multiple model training with different random seeds and select one of them as the base model. We visualize the JS divergence of the attention weight distribution between the base model and models trained with different random seeds for different methods. We conduct experiments on several test samples and each testing sample is divided into one of four bins by its maximum attention scores within the sentence. Here we use the RNN architecture.

We can see that Fig. \ref{fig:seeds_main} (c) and \ref{fig:seeds_main} (d) have heavy tails for the baseline vanilla attention on SST and Hate datasets. The violins covers more wider ranges along the x-axis. This can be interpreted as vanilla attention is unstable to random seeds. We can see from Fig. \ref{fig:seeds_main} (f)-(h) that while using our SEAT on SST, Hate and Rotten Tomatoes datasets, the violins are more narrow and their tails are lighter which imply SEAT is  more stable. This can be further confirmed by the fact  the violins of SEAL are more closer to zero than these of vanilla attention, which means their corresponds JSD values are more smaller. 

\paragraph{Embedding perturbation} We compare the stability of our SEAT with other baselines under embedding perturbation. In this setting, we mainly consider two metrics: JSD and TVD, which represents the  explainability stability and prediction stability respectively. Details of our setting are as follows.
\begin{enumerate}
     \item For testing the stability of explaination, it is sufficient to test the stability of the (modified) attention weights given by different methods. Specifically, we select one token of the input with embedding $x$, and then generate a perturbed embedding $x'=x + \mathcal{N}(0,\delta_x \mathbb{I})$ for some radius $\delta_x$. For each method, if we denote $w$ as its corresponding (modified) attention based on $x$ and $w'$ as its corresponding (modified) attention based on $x'$, we calculate $\text{JSD}(w, w')$ to measure such stability. 
     \item We also test the stability of prediction distribution. Specifically, similar to the above, we have two inputs, one is original and the other one is perturbed by some noise. For each method, we denote the prediction distribution of its attention based on the original input as   $y\in \mathbb{R}^{|\mathcal{Y}|}$, 
     and denote the prediction distribution of its attention based on the perturbed one as $y'$. Then we compute $\text{TVD}(y, y')$ to measure such stability.
\end{enumerate}
Results are shown in Tab. \ref{golden} and Tab. \ref{tab: more result main 3 metrics} in  Appendix. We can see that SEAT outperforms other baselines with RNN, BiLSTM and BERT under JSD and TVD evaluation metrics. Especially, we can see that the JSD of all our results are almost zero, which means SEAT is stable to perturbation for explaination. We can see also that the TVD for vanilla attention is large which means vanilla attention is extremely unstable to perturbation for its prediction distribution. However, the TVD of SEAT is small. 

\paragraph{Word perturbation}
We now aim to evaluate the stability of our proposed method under word perturbation.  Following \cite{yin2022sensitivity}, we select BERT as our main model in this part and conduct the perturbation in the following process: first we randomly choose $K$ words from a given sentence and then replace it with the closest synonyms. The distance of words are computed based on gensim~\cite{rehurek2011gensim}. We denote the original input and perturbed input as $x$ and $x'$ respectively. Then, similar to the above procedures, we can compute JSD and TVD for each method. Tab. \ref{tab_main_wordp} and Tab. \ref{tab_all_wordp} in  Appendix demonstrate that SEAT achieves SOTA for both JSD and TVD in this setting. Similar to the embedding perturbation case, we can see the JSD of SEAT is much smaller than it of the vanilla attention and it value is quite close to zero in all experiments, which indicates strong explaination stability against to word perturbation for SEAT. 

\begin{table}[thbp]
\centering
\resizebox{0.9\linewidth}{!}{
\begin{tabular}{cccccc} 
\toprule
\multirow{2}{*}{\textbf{Models}} & \multicolumn{2}{c}{\textbf{Ablation Setting}} & \multicolumn{3}{c}{\textbf{Metrics}}                       \\ 
\cmidrule(l){2-6}
                        & \bm{$\mathcal{L}_3$}     & \bm{$\mathcal{L}_{Topk}$}            & \textbf{Suff.$\downarrow$} & \textbf{TVD$\downarrow$} & \textbf{F1$\uparrow$}  \\ 
\midrule
\multirow{4}{*}{\textbf{RNN}}    &           &                          & 7.02E-04           & 21.464          & \textbf{0.814}         \\
                        & $\rightt$ &                          & 6.22E-04           & 1.966           & 0.804         \\
                        &           & $\rightt$                & 2.22E-04        & 2.997           & 0.782         \\
                        & $\rightt$ & $\rightt$                & \textbf{1.02E-04}        & \textbf{1.275}           & 0.813         \\ 
\midrule
\multirow{4}{*}{\textbf{BiLSTM}} &           &                          & 0.255           & 20.398          & \textbf{0.809}         \\
                        & $\rightt$ &                          & 0.016           & 1.214           & 0.802         \\
                        &           & $\rightt$                & 0.004           & 1.745           & 0.779         \\
                        & $\rightt$ & $\rightt$                & \textbf{4.37E-06}           & \textbf{1.095}           & 0.801         \\ 
\midrule
\multirow{4}{*}{\textbf{BERT}}   &           &                          & 0.310           & 2.617           & \textbf{0.912}         \\
                        & $\rightt$ &                          & 0.280           & 0.056           & 0.909         \\
                        &           & $\rightt$                & 0.090           & 0.157           & 0.907         \\
                        & $\rightt$ & $\rightt$                & \textbf{0.019}        & \textbf{0.028}           & 0.909         \\
\bottomrule
\end{tabular}
}
\caption{Ablation study of SEAT. We evaluate the effectiveness of $\mathcal{L}_3$ and  $\mathcal{L}_{Topk}$ in (\ref{eq:6}). 
Perturbation on the embedding space (radius $\delta_x=0.01$) are conducted on SST.}
\label{tab:ablation:main}
\vspace{-12pt}
\end{table}

\subsection{Evaluating Interpretability and Utility}
In this part, we measure the interpretability of SEAT and other baselines using comprehensiveness, sufficiency and sensitivity. Results are showed in Tab. \ref{golden-explanation} and Tab. \ref{tab: more result other three metrics} in Appendix. Our results show that SEAT outperforms other baselines on all three evaluation metrics with RNN, BiLSTM and BERT. This further confirms that enhancing stability in attention would derive a more faithful interpretation. Our SEAT improves the model interpretability.

In Tab. \ref{golden} and \ref{tab_main_wordp} we also compared the F1 score for different methods. We can see that while in some of the results, our method is not the best one. However, among these results, the difference between the best result and ours is quite small, which indicates that there is almost no accuracy deterioration in SEAT. Surprisingly, we can also see that SEAT is better than vanilla attention in most results and it could even achieve SOTA in some cases, such as when the model is RNN and the data is Emotion or Hate in Tab. \ref{golden}. 

\subsection{Ablation Study}
In ablation study, we evaluate each module (regularization) in (\ref{eq:6}). Specifically, we denote $\mathcal{L}_1$ as our main loss in (\ref{eq:1}), then we consider and evaluate different combinations by deleting $\mathcal{L}_{Topk}$ or/and $\mathcal{L}_{3}$, where $\mathcal{L}_3$ corresponds to the third term in (\ref{eq:6}). Note that if there is no $\mathcal{L}_{Topk}$ and $\mathcal{L}_3$, then the model will be the vanilla attention. 
Tab. \ref{tab:ablation:main},  Tab.  \ref{tab:ablation} and \ref{tab:ablation-2} in Appendix show that each regularizer in our objective function is indispensable and effective. Specifically, we can see the sufficiency will decrease significantly if we add the $\mathcal{L}_{Topk}$ loss. This is due to that 
$\mathcal{L}_{Topk}$ enforces a large  overlaps on top-k indices and thus it makes SEAT inherit the explainability of vanilla attention and it makes the model more stable. Moreover, in the case where there is $\mathcal{L}_{Topk}$, adding term $\mathcal{L}_3$ could further decrease sufficiency as it further improves stability. 

Since $\mathcal{L}_3$ is to make the prediction distribution stable against any randomness and perturbation, from Tab. \ref{tab:ablation} we can see adding this term could decrease the TVD, which means it improve the stability. Although $\mathcal{L}_{Topk}$ can also help to decrease TVD, we can see it is weaker than $\mathcal{L}_3$. Besides stability, $\mathcal{L}_3$ also can pull back the F1 score to make the model close to vanilla attention. We can see that in the case where there only exists $\mathcal{L}_{Topk}$, the F1 score decreases compared with vanilla attention. And adding $\mathcal{L}_3$ do help minimizing the gap. This is due to that $\mathcal{L}_3$ could improve the model generalization performance by making model more stable. 

\section{Conclusions}
In this paper, we provide a first rigorous definition namely SEAT as a substitute of attention to give a more faithful explanation. The definition has three properties: closeness of prediction, similarity of explainability and stability. We also propose a method to get such a SEAT, which could be considered as an ad hoc modification of the canonical attention. In experiments, we compare our SEAT with other methods using different metrics measuring explainability, performance and stability of prediction distribution and explainability. Results show that SEAT outperforms other baselines for all experiments on stability and explainability while it  almost has no accuracy degradation, which indicates SEAT could be considered as an effective and more faithful explanation tool.

\section*{Acknowledgements}
Di Wang and Lijie Hu were supported in part by the baseline funding BAS/1/1689-01-01, funding from the CRG grand URF/1/4663-01-01, FCC/1/1976-49-01 from CBRC and funding from the AI Initiative REI/1/4811-10-01 of King Abdullah University of Science and Technology (KAUST).  Di Wang was also supported by the funding of the SDAIA-KAUST Center of Excellence in Data Science and Artificial Intelligence (SDAIA-KAUST AI).

\bibliography{aaai23}

\begin{thebibliography}{30}
\providecommand{\natexlab}[1]{#1}

\bibitem[{Bahdanau, Cho, and Bengio(2015)}]{bahdanau2015neural}
Bahdanau, D.; Cho, K.; and Bengio, Y. 2015.
\newblock Neural Machine Translation by Jointly Learning to Align and
  Translate.
\newblock In Bengio, Y.; and LeCun, Y., eds., \emph{3rd International
  Conference on Learning Representations, {ICLR} 2015, San Diego, CA, USA, May
  7-9, 2015, Conference Track Proceedings}.

\bibitem[{Basile et~al.(2019)Basile, Bosco, Fersini, Nozza, Patti,
  Rangel~Pardo, Rosso, and Sanguinetti}]{basile-etal-2019-semeval}
Basile, V.; Bosco, C.; Fersini, E.; Nozza, D.; Patti, V.; Rangel~Pardo, F.~M.;
  Rosso, P.; and Sanguinetti, M. 2019.
\newblock {S}em{E}val-2019 Task 5: Multilingual Detection of Hate Speech
  Against Immigrants and Women in {T}witter.
\newblock In \emph{Proceedings of the 13th International Workshop on Semantic
  Evaluation}, 54--63. Minneapolis, Minnesota, USA: Association for
  Computational Linguistics.

\bibitem[{Choi et~al.(2016)Choi, Bahadori, Sun, Kulas, Schuetz, and
  Stewart}]{choi2016retain}
Choi, E.; Bahadori, M.~T.; Sun, J.; Kulas, J.; Schuetz, A.; and Stewart, W.~F.
  2016.
\newblock {RETAIN:} An Interpretable Predictive Model for Healthcare using
  Reverse Time Attention Mechanism.
\newblock In Lee, D.~D.; Sugiyama, M.; von Luxburg, U.; Guyon, I.; and Garnett,
  R., eds., \emph{Advances in Neural Information Processing Systems 29: Annual
  Conference on Neural Information Processing Systems 2016, December 5-10,
  2016, Barcelona, Spain}, 3504--3512.

\bibitem[{DeYoung et~al.(2020)DeYoung, Jain, Rajani, Lehman, Xiong, Socher, and
  Wallace}]{deyoung2020eraser}
DeYoung, J.; Jain, S.; Rajani, N.~F.; Lehman, E.; Xiong, C.; Socher, R.; and
  Wallace, B.~C. 2020.
\newblock ERASER: A Benchmark to Evaluate Rationalized NLP Models.
\newblock In \emph{Proceedings of the 58th Annual Meeting of the Association
  for Computational Linguistics}, 4443--4458.

\bibitem[{Dombrowski et~al.(2019)Dombrowski, Alber, Anders, Ackermann,
  M{\"{u}}ller, and Kessel}]{Dombrowski-etal19geometry}
Dombrowski, A.; Alber, M.; Anders, C.~J.; Ackermann, M.; M{\"{u}}ller, K.; and
  Kessel, P. 2019.
\newblock Explanations can be manipulated and geometry is to blame.
\newblock In Wallach, H.~M.; Larochelle, H.; Beygelzimer, A.;
  d'Alch{\'{e}}{-}Buc, F.; Fox, E.~B.; and Garnett, R., eds., \emph{Advances in
  Neural Information Processing Systems 32: Annual Conference on Neural
  Information Processing Systems 2019, NeurIPS 2019, December 8-14, 2019,
  Vancouver, BC, Canada}, 13567--13578.

\bibitem[{Dong et~al.(2019)Dong, Li, Rezagholizadeh, and
  Cheung}]{dong2019editnts}
Dong, Y.; Li, Z.; Rezagholizadeh, M.; and Cheung, J. C.~K. 2019.
\newblock {E}dit{NTS}: An Neural Programmer-Interpreter Model for Sentence
  Simplification through Explicit Editing.
\newblock In \emph{Proceedings of the 57th Annual Meeting of the Association
  for Computational Linguistics}, 3393--3402. Florence, Italy: Association for
  Computational Linguistics.

\bibitem[{Du, Liu, and Hu(2020)}]{du2019techniques}
Du, M.; Liu, N.; and Hu, X. 2020.
\newblock Techniques for interpretable machine learning.
\newblock \emph{Commun. {ACM}}, 63(1): 68--77.

\bibitem[{Galassi, Lippi, and Torroni(2020)}]{galassi2020attention}
Galassi, A.; Lippi, M.; and Torroni, P. 2020.
\newblock Attention in natural language processing.
\newblock \emph{IEEE Transactions on Neural Networks and Learning Systems},
  32(10): 4291--4308.

\bibitem[{Ghorbani, Abid, and Zou(2019)}]{ghorbani2019interpretation}
Ghorbani, A.; Abid, A.; and Zou, J.~Y. 2019.
\newblock Interpretation of Neural Networks Is Fragile.
\newblock In \emph{The Thirty-Third {AAAI} Conference on Artificial
  Intelligence, {AAAI} 2019, The Thirty-First Innovative Applications of
  Artificial Intelligence Conference, {IAAI} 2019, The Ninth {AAAI} Symposium
  on Educational Advances in Artificial Intelligence, {EAAI} 2019, Honolulu,
  Hawaii, USA, January 27 - February 1, 2019}, 3681--3688. {AAAI} Press.

\bibitem[{Jacovi and Goldberg(2020)}]{jacovi2020towards}
Jacovi, A.; and Goldberg, Y. 2020.
\newblock Towards Faithfully Interpretable NLP Systems: How should we define
  and evaluate faithfulness?
\newblock \emph{arXiv preprint arXiv:2004.03685}.

\bibitem[{Jain and Wallace(2019)}]{jain2019attention}
Jain, S.; and Wallace, B.~C. 2019.
\newblock {A}ttention is not {E}xplanation.
\newblock In \emph{Proceedings of the 2019 Conference of the North {A}merican
  Chapter of the Association for Computational Linguistics: Human Language
  Technologies, Volume 1 (Long and Short Papers)}, 3543--3556. Minneapolis,
  Minnesota: Association for Computational Linguistics.

\bibitem[{Kitada and Iyatomi(2021)}]{kitada2021attention}
Kitada, S.; and Iyatomi, H. 2021.
\newblock Attention Meets Perturbations: Robust and Interpretable Attention
  With Adversarial Training.
\newblock \emph{{IEEE} Access}, 9: 92974--92985.

\bibitem[{Lei(2017)}]{lei2017interpretable}
Lei, T. 2017.
\newblock \emph{Interpretable neural models for natural language processing}.
\newblock Ph.D. thesis, Massachusetts Institute of Technology, Cambridge,
  {USA}.

\bibitem[{Maas et~al.(2011)Maas, Daly, Pham, Huang, Ng, and
  Potts}]{maas-EtAl:2011:ACL-HLT2011}
Maas, A.~L.; Daly, R.~E.; Pham, P.~T.; Huang, D.; Ng, A.~Y.; and Potts, C.
  2011.
\newblock Learning Word Vectors for Sentiment Analysis.
\newblock In \emph{Proceedings of the 49th Annual Meeting of the Association
  for Computational Linguistics: Human Language Technologies}, 142--150.
  Portland, Oregon, USA: Association for Computational Linguistics.

\bibitem[{Madry et~al.(2018)Madry, Makelov, Schmidt, Tsipras, and
  Vladu}]{madry2018towards}
Madry, A.; Makelov, A.; Schmidt, L.; Tsipras, D.; and Vladu, A. 2018.
\newblock Towards Deep Learning Models Resistant to Adversarial Attacks.
\newblock In \emph{6th International Conference on Learning Representations,
  {ICLR} 2018, Vancouver, BC, Canada, April 30 - May 3, 2018, Conference Track
  Proceedings}. OpenReview.net.

\bibitem[{Martins and Astudillo(2016)}]{martins2016softmax}
Martins, A. F.~T.; and Astudillo, R.~F. 2016.
\newblock From Softmax to Sparsemax: {A} Sparse Model of Attention and
  Multi-Label Classification.
\newblock In Balcan, M.; and Weinberger, K.~Q., eds., \emph{Proceedings of the
  33nd International Conference on Machine Learning, {ICML} 2016, New York
  City, NY, USA, June 19-24, 2016}, volume~48 of \emph{{JMLR} Workshop and
  Conference Proceedings}, 1614--1623. JMLR.org.

\bibitem[{Miyato, Dai, and Goodfellow(2016)}]{miyato2016adversarial}
Miyato, T.; Dai, A.~M.; and Goodfellow, I. 2016.
\newblock Adversarial training methods for semi-supervised text classification.
\newblock \emph{arXiv preprint arXiv:1605.07725}.

\bibitem[{Mohammad et~al.(2018)Mohammad, Bravo-Marquez, Salameh, and
  Kiritchenko}]{emotion}
Mohammad, S.; Bravo-Marquez, F.; Salameh, M.; and Kiritchenko, S. 2018.
\newblock Semeval-2018 task 1: Affect in tweets.
\newblock In \emph{Proceedings of the 12th international workshop on semantic
  evaluation}, 1--17.

\bibitem[{Mohankumar et~al.(2020)Mohankumar, Nema, Narasimhan, Khapra,
  Srinivasan, and Ravindran}]{mohankumar2020towards}
Mohankumar, A.~K.; Nema, P.; Narasimhan, S.; Khapra, M.~M.; Srinivasan, B.~V.;
  and Ravindran, B. 2020.
\newblock Towards Transparent and Explainable Attention Models.
\newblock In \emph{Proceedings of the 58th Annual Meeting of the Association
  for Computational Linguistics}, 4206--4216.

\bibitem[{Otter, Medina, and Kalita(2021)}]{otter2020survey}
Otter, D.~W.; Medina, J.~R.; and Kalita, J.~K. 2021.
\newblock A Survey of the Usages of Deep Learning for Natural Language
  Processing.
\newblock \emph{{IEEE} Trans. Neural Networks Learn. Syst.}, 32(2): 604--624.

\bibitem[{Pang and Lee(2005)}]{pang2005seeing}
Pang, B.; and Lee, L. 2005.
\newblock Seeing Stars: Exploiting Class Relationships for Sentiment
  Categorization with Respect to Rating Scales.
\newblock In \emph{Proceedings of the 43rd Annual Meeting of the Association
  for Computational Linguistics (ACL’05)}, 115--124.

\bibitem[{Rehurek and Sojka(2011)}]{rehurek2011gensim}
Rehurek, R.; and Sojka, P. 2011.
\newblock Gensim--python framework for vector space modelling.
\newblock \emph{NLP Centre, Faculty of Informatics, Masaryk University, Brno,
  Czech Republic}, 3(2).

\bibitem[{Ribeiro, Singh, and Guestrin(2016)}]{ribeiro2016should}
Ribeiro, M.; Singh, S.; and Guestrin, C. 2016.
\newblock {``}Why Should {I} Trust You?{''}: Explaining the Predictions of Any
  Classifier.
\newblock In \emph{Proceedings of the 2016 Conference of the North {A}merican
  Chapter of the Association for Computational Linguistics: Demonstrations},
  97--101. San Diego, California: Association for Computational Linguistics.

\bibitem[{Sato et~al.(2018)Sato, Suzuki, Shindo, and
  Matsumoto}]{sato2018interpretable}
Sato, M.; Suzuki, J.; Shindo, H.; and Matsumoto, Y. 2018.
\newblock Interpretable adversarial perturbation in input embedding space for
  text.
\newblock In \emph{Proceedings of the 27th International Joint Conference on
  Artificial Intelligence}, 4323--4330.

\bibitem[{Serrano and Smith(2019)}]{serrano2019attention}
Serrano, S.; and Smith, N.~A. 2019.
\newblock Is Attention Interpretable?
\newblock In \emph{Proceedings of the 57th Annual Meeting of the Association
  for Computational Linguistics}, 2931--2951. Florence, Italy: Association for
  Computational Linguistics.

\bibitem[{Socher et~al.(2013)Socher, Perelygin, Wu, Chuang, Manning, Ng, and
  Potts}]{socher2013recursive}
Socher, R.; Perelygin, A.; Wu, J.; Chuang, J.; Manning, C.~D.; Ng, A.~Y.; and
  Potts, C. 2013.
\newblock Recursive deep models for semantic compositionality over a sentiment
  treebank.
\newblock In \emph{Proceedings of the 2013 conference on empirical methods in
  natural language processing}, 1631--1642.

\bibitem[{Vaswani et~al.(2017)Vaswani, Shazeer, Parmar, Uszkoreit, Jones,
  Gomez, Kaiser, and Polosukhin}]{vaswani2017attention}
Vaswani, A.; Shazeer, N.; Parmar, N.; Uszkoreit, J.; Jones, L.; Gomez, A.~N.;
  Kaiser, L.; and Polosukhin, I. 2017.
\newblock Attention is All you Need.
\newblock In Guyon, I.; von Luxburg, U.; Bengio, S.; Wallach, H.~M.; Fergus,
  R.; Vishwanathan, S. V.~N.; and Garnett, R., eds., \emph{Advances in Neural
  Information Processing Systems 30: Annual Conference on Neural Information
  Processing Systems 2017, December 4-9, 2017, Long Beach, CA, {USA}},
  5998--6008.

\bibitem[{Wiegreffe and Pinter(2019)}]{wiegreffe2019attention}
Wiegreffe, S.; and Pinter, Y. 2019.
\newblock Attention is not not Explanation.
\newblock In \emph{Proceedings of the 2019 Conference on Empirical Methods in
  Natural Language Processing and the 9th International Joint Conference on
  Natural Language Processing (EMNLP-IJCNLP)}, 11--20. Hong Kong, China:
  Association for Computational Linguistics.

\bibitem[{Yeh et~al.(2019)Yeh, Hsieh, Suggala, Inouye, and
  Ravikumar}]{yeh2019fidelity}
Yeh, C.; Hsieh, C.; Suggala, A.~S.; Inouye, D.~I.; and Ravikumar, P. 2019.
\newblock On the (In)fidelity and Sensitivity of Explanations.
\newblock In Wallach, H.~M.; Larochelle, H.; Beygelzimer, A.;
  d'Alch{\'{e}}{-}Buc, F.; Fox, E.~B.; and Garnett, R., eds., \emph{Advances in
  Neural Information Processing Systems 32: Annual Conference on Neural
  Information Processing Systems 2019, NeurIPS 2019, December 8-14, 2019,
  Vancouver, BC, Canada}, 10965--10976.

\bibitem[{Yin et~al.(2022)Yin, Shi, Hsieh, and Chang}]{yin2022sensitivity}
Yin, F.; Shi, Z.; Hsieh, C.-J.; and Chang, K.-W. 2022.
\newblock On the Sensitivity and Stability of Model Interpretations in {NLP}.
\newblock In \emph{Proceedings of the 60th Annual Meeting of the Association
  for Computational Linguistics (Volume 1: Long Papers)}, 2631--2647. Dublin,
  Ireland: Association for Computational Linguistics.

\end{thebibliography}

\clearpage
\appendix

\section{Implementation Details}

\begin{table}[h]
    \centering
    \small
    \resizebox{1.0\linewidth}{!}{
    \begin{tabular}{cccc} 
\toprule
\multirow{2}{*}{\textbf{Dataset}} & \textbf{Avg. Length} & \textbf{Train Size} & \textbf{Test Size}  \\ 
\cmidrule{2-4}
                         & (texts)     & (neg/pos)  & (neg/pos)  \\ 
\midrule
Emotion~                 & 91          & 1400/708   & 558/358    \\
SST~                     & 19          & 3610/3310  & 909/912    \\
Hate~                    & 23          & 3783/5217  & 1252/1718  \\
Rotten Tomatoes~         & 23          & 4265/4265  & 533/533    \\
\bottomrule
\end{tabular}
    }
    \caption{Dataset statistics. \label{tab:data}}
\end{table}

\subsection{Experimental Setup}
\paragraph{Dataset Details} Following the previous study~\cite{serrano2019attention,wiegreffe2019attention}, we mainly focus on the experimental evaluation of binary classification tasks in this paper. To be specific, we conduct experiments on the following datasets: Stanford Sentiment Treebank (SST)~\cite{socher2013recursive}, Emotion Recognition~\cite{emotion}, Hate~\cite{basile-etal-2019-semeval} and Rotten Tomatoes~\cite{pang2005seeing}. For each dataset, we extract the samples with label 0 or 1 in the dataset. The details of datasets are presented in \ref{tab:data}. We use the default train-test split configuration offered by the Huggingface for each dataset. Note our framework is task-agnostic and we will leave examining our framework for other NLP tasks such as question answering (QA) and natural language inference (NLI) as future work. 

\paragraph{Models and architectures} We mainly study encoder-decoder architectures in this paper following ~\cite{wiegreffe2019attention}. For the encoder, we consider three kinds of networks as feature extractors: RNN, BiLSTM, and BERT. For the decoder, we apply one simple MLP following with a tanh-attention layer ~\cite{bahdanau2015neural} and a softmax layer ~\cite{vaswani2017attention}, which is followed by ~\cite{jain2019attention}.

\paragraph{Metrics} In our experiments, we seek to evaluate stability, explainability and performance of our model.

For evaluating the stability. Following~\cite{jain2019attention}, for each method, we use Jensen-Shannon Divergence (JSD) between its attention with on perturbation and its attention under perturbation to evaluate the {\bf stability of explainablity} of the learned attention, which is defined as $$\text{JSD}(\alpha_1, \alpha_2) = \frac{1}{2} \text{KL}[\alpha_1||\frac{\alpha_1 + \alpha_2}{2}] + \frac{1}{2}\text{KL}[\alpha_2||\frac{\alpha_1 + \alpha_2}{2}],$$ where $KL$ is the $KL$ divergence between two  distributions.  We also the Total
Variation Distance (TVD) between the prediction distribution with no perturbation and the prediction distribution under perturbation is used to measure {\bf prediction stability},  which can be defined as, $$\text{TVD}(\hat{y}_1, \hat{y}_2) = \frac{1}{2}\sum_{i=1}^{|\mathcal{Y}|} |\hat{y}_{1i}-\hat{y}_{2i}|.$$ To evaluate the performance of our model performance, we report the F1 score of models on clean test set.

To evaluate the model interpretability, we leverage three different metrics in the previous study~\cite{yin2022sensitivity,deyoung2020eraser}. Specifically, we use \textit{comprehensiveness, sufficiency and sensitivity} to evaluate the model performance on interpretation and stability. Following \cite{deyoung2020eraser}, for given input sentence $x_i$, we identify the rationales $r_i$ (i.e., sequence of tokens that help classifying) of this sentence by exacting the top-$k$ elements based on attention values. Note that for the consideration of computation cost, we sample certain percentage of data to evaluate comprehensiveness and sufficiency. 

\paragraph{Comprehensiveness, Sufficiency \cite{deyoung2020eraser}. }
Following the notation in the previous study, we denote the model as $m(\cdot)$.  Intuitively, model with higher interpretation ability should be less confident in making decision once the rationales are removed. Formally, we have
\begin{equation}
    \text { comprehensiveness }=m\left(x_{i}\right)_{j}-m\left(x_{i} \backslash r_{i}\right)_{j}
\end{equation}
Another related metric of comprehensiveness is  sufficiency. It captures the degree that extracted rationales are sufficient for the model to make classification. Formally, we have
\begin{equation}
    \text { sufficiency }=m\left(x_{i}\right)_{j}-m\left(r_{i}\right)_{j}
\end{equation}

\paragraph{Sensitivity. \cite{yin2022sensitivity}}
As point out in \cite{yin2022sensitivity}, models with higher interpretability should have lower sensitivity under some local and adversarial perturbations on important tokens. Specifically, given a sequence of rationales $r_k$, they evaluate the sensitivity by only adding perturbations to its corresponding embedding $e\left(r_k\right)$ and keeps the other token embedding the same as original. Then, they measures the minimal perturbation norm, denoted as $\epsilon_{r_k}$, that changes the model prediction for this instance:
\begin{equation}
\epsilon_{r_{k}}=\min \left\|\boldsymbol{\delta}_{\boldsymbol{r}_{\boldsymbol{k}}}\right\|_{F} \quad \text { s.t. } \quad f\left(e(x)+\boldsymbol{\delta}_{\boldsymbol{r}_{\boldsymbol{k}}}\right) \neq y
\end{equation}
where $\norm{\cdot}_F$ is the Frobenius norm of the matrix, and $\boldsymbol{\delta_{r_k}} \in \mathcal{R}^{n \times d}$ denotes the perturbation matrix where only the columns for tokens in $r_k$ have non-zero elements. Given that the exact computation of $\epsilon_{r_k}$ is intractable, they use the PGD attack \citep{madry2018towards} with a binary search to approximate $\epsilon_{r_k}$. Following \cite{yin2022sensitivity}, in practice, we vary the size of $r_k$, compute multiple $\epsilon_{r_k}$, and summarize them with the area under the curve (AUC) score. 
\subsection{Baselines and Implementation}
We compare our method with six baseline methods, which are the only works that study the self-explaining interpretation and stability of the attention mechanism. We first briefly introduce the consider baseline methods for completeness. 
\begin{itemize}
    \item \textit{Vanilla}~\cite{wiegreffe2019attention}: model with an attention layer trained with natural cross entropy loss $\ell_\theta=\sum_i \text{CE} (\hat{y}(x_i),y_i)$. The cross-entropy of the distribution $q$ relative to a distribution $p$ over a given set is defined as follows:
    \begin{equation*}
        H(p,q) = - \mathbb{E}_p[\log q],
    \end{equation*}
    where $\mathbb{E}_p[\cdot]$ is the expected value operator with respect to the distribution $p$. 
    \item \textit{Word AT}~\cite{miyato2016adversarial}: model trained with weighted sum of benign loss and adversarial loss, i.e., $\ell_\theta=\sum_i \text{CE} (\hat{y}(x_i),y_i)+\lambda\text{CE} (\hat{y}(\tilde{x}_i),y_i)$ where the perturbation is conducted on the token embedding. 
     \item \textit{Word iAT}~\cite{sato2018interpretable}: similar to Word-AT method, Word-iAT conducts adversarial training on the token embedding space. The difference is that the adversarial perturbation constructed by Word-AT on each word (token) is based on the ``interpretable'' direction set. Specifically, given the  embedding $e(x)$ of all the words $x$ in the dictionary $D$ and a word embedding $e(x_i)$ that aims to be perturbed, the perturbation direction $\Delta e(x_i)$ must be the linear combination of vector sampled from $S_{\Delta e(x_i)}=\{\forall j \neq i, \Delta e(x_{ij})|\Delta e(x_{ij})=e(x_j)-e(x_i)\}$, i.e., $\Delta e(x_i)=\sum_j\alpha_{ij}\Delta e(x_{ij})$, where $\alpha_j \in \mathbb{R}$ and $\Delta e(x_{ij})\in S_{\Delta e(x_i)}$
    \item \textit{Attention-AT, Attention-iAT, Attention-RP}\cite{kitada2021attention}: The main ideas of the first two methods are similar as Word-AT and Word-iAT respectively. The difference is that here we conduct the adversarial training on the attention weights. Note that for the Attention-iAT method, the perturbation are constructed based on the difference of attention weight within sentence, i.e., $\Delta a_i=\sum_j\gamma_{ij}(a_j-a_i)$, where $\gamma_{ij} \in \mathbb{R}$. As for the Attention-RP method, model is trained with random noise perturbations on attention weights. 
\end{itemize}
All the baseline methods are implemented on our own based on the guidance of original paper and official code (if it exists). 
\subsection{Model Training and Hyper-parameter}
All models are implemented based on the PyTorch \footnote[1]{https://pytorch.org/} library. All experiments are conducted on NVIDIA RTX A5000 GPUs. For RNN and BiLSTM, the initial learning rate, training epoch, and batch size is set as 0.01, 20 and 32 respectively. For RNN and BiLSTM classifier, we use an one-layer RNN/BiLSTM encoder with hidden size of 128. The embedding of input token is initialized with the 300-dimensional pre-trained \texttt{fasttext} word embedding. For BERT, we use the bert-base-uncased model. We fine-tune BERT model on each dataset, using the initial learning rate of 5e-5, batch size 8, and training epoch 4. Adam are used as optimizer for all models. Early stopping on testing F1 metric is conducted during training to prevent over-fitting. In each of experiment, we use the following hyper-parameter setting by default. As for the coefficient of loss objective in Eq.\ref{eq:4}, $\lambda_1=1$ and $\lambda_2=1000$. As for the top-$k$ overlap surrogate loss, we set the $k$ as 7. 

\begin{table}[!ht]
\centering
\resizebox{1.0\linewidth}{!}{
\begin{tabular}{cccccc} 
\toprule
\multirow{2}{*}{\textbf{Models}} & \multicolumn{2}{c}{\textbf{Ablation Setting}} & \multicolumn{3}{c}{\textbf{Metrics}}                       \\ 
\cmidrule(l){2-6}
                        & \bm{$\mathcal{L}_3$}     & \bm{$\mathcal{L}_{Topk}$}            & \textbf{JSD$\downarrow$} & \textbf{TVD$\downarrow$} & \textbf{F1$\uparrow$}  \\ 
\midrule
\multirow{4}{*}{\textbf{RNN}}    &           &                          & 0.025           & 21.464          & \textbf{0.814}         \\
                        & $\rightt$ &                          & 0.017           & 1.966           & 0.804         \\
                        &           & $\rightt$                & 8.21E-08        & 2.997           & 0.782         \\
                        & $\rightt$ & $\rightt$                & \textbf{6.98E-08}        & \textbf{1.275}           & 0.813         \\ 
\midrule
\multirow{4}{*}{\textbf{BiLSTM}} &           &                          & 0.059           & 20.398          & \textbf{0.809}         \\
                        & $\rightt$ &                          & 0.033           & 1.214           & 0.802         \\
                        &           & $\rightt$                & 0.030           & 1.745           & 0.779         \\
                        & $\rightt$ & $\rightt$                & \textbf{0.028}           & \textbf{1.095}           & 0.801         \\ 
\midrule
\multirow{4}{*}{\textbf{BERT}}   &           &                          & 0.140           & 2.617           & \textbf{0.912}         \\
                        & $\rightt$ &                          & 0.114           & 0.056           & 0.909         \\
                        &           & $\rightt$                & 0.006           & 0.157           & 0.907         \\
                        & $\rightt$ & $\rightt$                & \textbf{2.49E-07}        & \textbf{0.028}           & 0.909         \\
\bottomrule
\end{tabular}
}
\caption{Ablation study of the proposed method. We evaluate the effectiveness of the third term $\mathcal{L}_3$ and the second term $\mathcal{L}_{Topk}$ in (\ref{eq:6}) on three different architectures. JSD, TVD and F1 are reported to measure the stability and performance of the model. Here we conduct perturbation on the embedding space with radius of 0.01 on SST dataset.}
\label{tab:ablation}
\end{table}

\begin{table}[htbp]
\centering
\resizebox{1.0\linewidth}{!}{
\begin{tabular}{cccccc} 
\toprule
\multirow{2}{*}{\textbf{Models}} & \multicolumn{2}{c}{\textbf{Ablation Setting}} & \multicolumn{3}{c}{\textbf{Metrics}}                                                                                \\ 
\cmidrule(l){2-6}
                                 & $\mathcal{L}_3$ & $\mathcal{L}_{Topk}$        & \textbf{Comp.$\uparrow$}                                 & \textbf{Suff.$\downarrow$} & \textbf{Sens.$\downarrow$}  \\ 
\midrule
\multirow{4}{*}{\textbf{RNN}}    &                 &                             & 5.744                                                    & 7.02E-04                   & 0.090                       \\
                                 & $\rightt$       &                             & 5.280                                                    & 6.22E-04                   & 0.081                       \\
                                 &                 & $\rightt$                   & 5.944                                                    & 2.22E-04                   & 0.088                       \\
                                 & $\rightt$       & $\rightt$                   & \textbf{6.016}                                           & \textbf{1.02E-04}          & \textbf{0.076}              \\ 
\midrule
\multirow{4}{*}{\textbf{BiLSTM}} &                 &                             & 5.182                                                    & 0.255                      & 0.098                       \\
                                 & $\rightt$       &                             & 2.219                                                    & 0.016                      & 0.087                       \\
                                 &                 & $\rightt$                   & 5.167                                                    & 0.004                      & 0.095                       \\
                                 & $\rightt$       & $\rightt$                   & \begin{tabular}[c]{@{}c@{}}\textbf{5.435}\\\end{tabular} & \textbf{4.37E-06}          & \textbf{0.076}              \\ 
\midrule
\multirow{4}{*}{\textbf{BERT}}   &                 &                             & 0.003                                                    & 0.310                      & 0.009                       \\
                                 & $\rightt$       &                             & 0.002                                                    & 0.280                      & 0.005                       \\
                                 &                 & $\rightt$                   & 0.210                                                    & 0.090                      & 0.006                       \\
                                 & $\rightt$       & $\rightt$                   & \textbf{0.497}                                           & \textbf{0.019}             & \textbf{0.004}              \\
\bottomrule
\end{tabular}
}
\caption{Ablation study of the proposed method on the explainability. We evaluate the effectiveness of the  the second term $\mathcal{L}_{Topk}$ and third term $\mathcal{L}_3$ in (\ref{eq:6}) on three different architectures. Comprehensiveness, Sufficiency and Sensitivity are reported to measure the explainability of the model. Here we conduct perturbation on the embedding space with radius of 0.01 on SST dataset.}
\label{tab:ablation-2}
\end{table}

\section{Additional Experimental Results}
\subsection{Surrogate Top-$k$ Loss Performance}
\begin{figure*}[!thbp]
\centering
\includegraphics[scale=0.23]{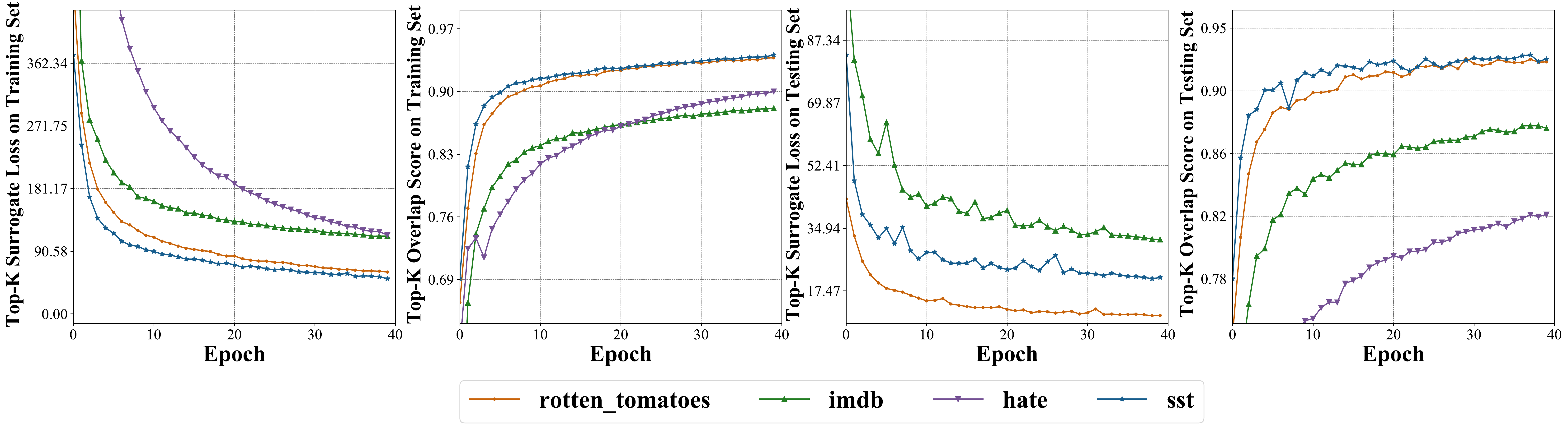}
\caption{Surrogate Top-$k$ Loss Performance. Evaluation curves of Surrogate Top-$k$ loss performance. The left two figures are the training loss and overlapping score on training set. The right two figures are the testing loss and overlapping score on testing set. The X-axis is the number of epoch. The Y-axis is the performance under the setting. Results show that our purposed Top-$k$ overlap surrogate loss is a good approximation of true non-differentiable Top-$k$ loss. 
}
\label{fig:topk}
\end{figure*}

In this part, we test the efficiency of our surrogate top-$k$ loss function. As we can see from the Fig.\ref{fig:topk}, the top-$k$ loss is decreasing along with the epoch number in all datasets. And the overlapping rate reaches 93 percent for SST and Rotten Tomatoes datasets. Results show that our surrogate top-$k$ loss is valid to approximate the true top-$k$ overlap score.

\subsection{More Results on Stability and Explanability of Attention}
In this section, we show more results on the stability and explanability of different methods with different perturbation radius $\delta$ on RNN, BiLSTM and BERT in Tab.\ref{tab: more result main 3 metrics} and Tab.\ref{tab: more result other three metrics}. The perturbation is conducted on the embedding space. 
\begin{table*}[ht]
\small
\centering
\label{tab:diff_rad}
\resizebox{1.0\textwidth}{!}{
\begin{tabular}{c|cccccccccccccc} 
\toprule
\multicolumn{1}{c}{\multirow{2}{*}{Radius($\delta_x$)}} & \multirow{2}{*}{Model}  & \multirow{2}{*}{Method} & \multicolumn{3}{c}{Emotion}                                                             & \multicolumn{3}{c}{SST}   & \multicolumn{3}{c}{Hate}  & \multicolumn{3}{c}{RottenT}   \\ 
\cmidrule(r){4-6}\cmidrule(r){7-9}\cmidrule(r){10-12}\cmidrule(lr){13-15}
\multicolumn{1}{c}{}                                                        &                         &                         & JSD$\downarrow$ & TVD$\downarrow$ & F1$\uparrow$ & JSD      & TVD    & F1    & JSD      & TVD    & F1    & JSD       & TVD      & F1     \\ 
\midrule
\multirow{21}{*}{0.005}                                                     & \multirow{7}{*}{RNN}    & Vanilla                 & 0.002                        & 20.120                       & 0.677                     & 0.019    & 19.916 & 0.767 & 0.009    & 15.627 & 0.513 & 0.008     & 19.240   & 0.763  \\
                                                                            &                         & Word-AT                 & 0.016                        & 1.634                        & 0.642                     & 0.017    & 2.186  & 0.800 & 0.035    & 1.618  & 0.519 & 0.058     & 1.401    & 0.768  \\
                                                                            &                         & Word-iAT                & 0.027                        & 1.570                        & 0.637                     & 0.021    & 1.331  & 0.811 & 0.025    & 1.387  & 0.537 & 0.092     & 1.719    & 0.759  \\
                                                                            &                         & Attention-RP            & 0.026                        & 3.281                        & 0.671                     & 0.028    & 2.209  & 0.792 & 0.025    & 2.650  & 0.554 & 0.009     & 3.759    & 0.770  \\
                                                                            &                         & Attention-AT            & 0.055                        & 2.738                        & 0.665                     & 0.047    & 2.595  & 0.782 & 0.032    & 2.195  & 0.528 & 0.068     & 4.328    & 0.755  \\
                                                                            &                         & Attention-iAT           & 0.017                        & 3.573                        & 0.645                     & 0.050    & 2.821  & 0.746 & 0.039    & 2.261  & 0.533 & 0.055     & 1.645    & 0.753  \\
                                                                            &                         & SEAT(\textbf{Ours})     & 8.41E-09                     & 1.320                        & 0.689                     & 4.19E-08 & 1.143  & 0.803 & 2.51E-09 & 1.263  & 0.579 & -2.87E-10 & 1.174    & 0.762  \\ 
\cmidrule{2-15}
                                                                            & \multirow{7}{*}{BiLSTM} & Vanilla                 & 0.002                        & 23.440                       & 0.653                     & 0.037    & 19.215 & 0.809 & 0.062    & 15.805 & 0.524 & 0.009     & 20.197   & 0.764  \\
                                                                            &                         & Word-AT                 & 0.053                        & 1.632                        & 0.655                     & 0.025    & 1.080  & 0.794 & 0.058    & 2.126  & 0.528 & 0.039     & 1.223    & 0.772  \\
                                                                            &                         & Word-iAT                & 0.054                        & 1.902                        & 0.643                     & 0.029    & 1.120  & 0.809 & 0.078    & 1.899  & 0.525 & 0.064     & 1.187    & 0.750  \\
                                                                            &                         & Attention-RP            & 0.031                        & 1.391                        & 0.642                     & 0.036    & 1.396  & 0.772 & 0.054    & 1.388  & 0.522 & 0.068     & 1.539    & 0.764  \\
                                                                            &                         & Attention-AT            & 0.079                        & 1.617                        & 0.672                     & 0.030    & 1.748  & 0.779 & 0.058    & 1.582  & 0.523 & 0.080     & 2.129    & 0.766  \\
                                                                            &                         & Attention-iAT           & 0.035                        & 1.371                        & 0.651                     & 0.036    & 1.722  & 0.801 & 0.063    & 2.310  & 0.525 & 0.079     & 1.931    & 0.777  \\
                                                                            &                         & SEAT(\textbf{Ours})     & 9.68E-09                     & 0.736                        & 0.671                     & 2.70E-08 & 1.040  & 0.805 & 1.47E-08 & 1.081  & 0.543 & 2.06E-08  & 0.991    & 0.770  \\ 
\cmidrule{2-15}
                                                                            & \multirow{7}{*}{BERT}   & Vanilla                 & 0.026                        & 2.130                        & 0.721                     & 0.004    & 2.610  & 0.912 & 0.039    & 1.771  & 0.493 & 0.010     & 2.504    & 0.845  \\
                                                                            &                         & Word-AT                 & 0.013                        & 0.024                        & 0.694                     & 0.100    & 0.035  & 0.891 & 0.006    & 0.074  & 0.579 & 0.563     & 0.083    & 0.845  \\
                                                                            &                         & Word-iAT                & 0.456                        & 0.059                        & 0.658                     & 0.213    & 0.094  & 0.912 & 0.331    & 0.043  & 0.501 & 0.488     & 0.048    & 0.852  \\
                                                                            &                         & Attention-RP            & 0.039                        & 0.235                        & 0.657                     & 0.089    & 0.128  & 0.893 & 0.085    & 0.278  & 0.554 & 0.078     & 0.143    & 0.817  \\
                                                                            &                         & Attention-AT            & 0.081                        & 0.019                        & 0.707                     & 0.006    & 0.157  & 0.907 & 0.035    & 0.230  & 0.510 & 0.049     & 0.193    & 0.818  \\
                                                                            &                         & Attention-iAT           & 0.126                        & 0.228                        & 0.684                     & 0.147    & 0.204  & 0.915 & 0.080    & 0.271  & 0.512 & 0.135     & 0.187    & 0.831  \\
                                                                            &                         & SEAT(\textbf{Ours})     & 1.70E-08                     & 0.001                        & 0.717                     & 1.29E-07 & 0.028  & 0.911 & 6.47E-06 & 0.003  & 0.575 & 2.61E-05  & 0.033    & 0.848  \\ 
\midrule
\multirow{21}{*}{0.01}                                                      & \multirow{7}{*}{RNN}    & Vanilla                 & 0.002                        & 20.119                       & 0.657                     & 0.021    & 20.716 & 0.797 & 0.010    & 15.798 & 0.513 & 0.008     & 19.383   & 0.763  \\
                                                                            &                         & Word-AT                 & 0.017                        & 1.385                        & 0.629                     & 0.017    & 1.266  & 0.814 & 0.023    & 1.238  & 0.532 & 0.054     & 1.196    & 0.765  \\
                                                                            &                         & Word-iAT                & 0.027                        & 1.576                        & 0.637                     & 0.021    & 1.331  & 0.811 & 0.025    & 1.248  & 0.537 & 0.092     & 1.716    & 0.759  \\
                                                                            &                         & Attention-RP            & 0.026                        & 3.296                        & 0.671                     & 0.028    & 2.208  & 0.792 & 0.025    & 2.662  & 0.554 & 0.009     & 3.750    & 0.770  \\
                                                                            &                         & Attention-AT            & 0.055                        & 2.717                        & 0.665                     & 0.048    & 2.587  & 0.782 & 0.032    & 2.209  & 0.528 & 0.068     & 4.358    & 0.755  \\
                                                                            &                         & Attention-iAT           & 0.017                        & 3.554                        & 0.645                     & 0.050    & 2.837  & 0.746 & 0.039    & 2.261  & 0.533 & 0.055     & 1.646    & 0.753  \\
                                                                            &                         & SEAT(\textbf{Ours})     & 1.40E-08                     & 1.250                        & 0.688                     & 6.44E-08 & 1.175  & 0.813 & 6.48E-10 & 1.198  & 0.579 & 1.16E-07  & 0.975    & 0.762  \\ 
\cmidrule{2-15}
                                                                            & \multirow{7}{*}{BiLSTM} & Vanilla                 & 0.002                        & 23.432                       & 0.643                     & 0.059    & 20.398 & 0.809 & 0.072    & 16.312 & 0.524 & 0.010     & 20.322   & 0.764  \\
                                                                            &                         & Word-AT                 & 0.045                        & 1.551                        & 0.658                     & 0.033    & 1.214  & 0.802 & 0.071    & 1.609  & 0.516 & 0.062     & 1.214    & 0.765  \\
                                                                            &                         & Word-iAT                & 0.054                        & 1.899                        & 0.643                     & 0.029    & 1.594  & 0.809 & 0.078    & 1.899  & 0.525 & 0.064     & 1.203    & 0.750  \\
                                                                            &                         & Attention-RP            & 0.031                        & 1.390                        & 0.642                     & 0.036    & 1.398  & 0.772 & 0.054    & 1.386  & 0.522 & 0.068     & 1.540    & 0.764  \\
                                                                            &                         & Attention-AT            & 0.080                        & 1.613                        & 0.672                     & 0.030    & 1.745  & 0.779 & 0.058    & 1.581  & 0.523 & 0.080     & 2.130    & 0.766  \\
                                                                            &                         & Attention-iAT           & 0.035                        & 1.370                        & 0.651                     & 0.036    & 1.724  & 0.801 & 0.063    & 2.315  & 0.525 & 0.079     & 1.932    & 0.777  \\
                                                                            &                         & SEAT(\textbf{Ours})     & 1.04E-08                     & 0.960                        & 0.664                     & 1.63E-07 & 1.116  & 0.804 & 8.17E-09 & 1.167  & 0.547 & 6.44E-09  & 0.948    & 0.770  \\ 
\cmidrule{2-15}
                                                                            & \multirow{7}{*}{BERT}   & Vanilla                 & 0.033                        & 2.134                        & 0.721                     & 0.004    & 2.617  & 0.912 & 0.047    & 1.772  & 0.493 & 0.010     & 2.508    & 0.845  \\
                                                                            &                         & Word-AT                 & 0.004                        & 0.040                        & 0.694                     & 0.118    & 0.046  & 0.909 & 0.309    & 0.074  & 0.540 & 0.390     & 0.086    & 0.829  \\
                                                                            &                         & Word-iAT                & 0.456                        & 0.059                        & 0.658                     & 0.213    & 0.107  & 0.912 & 0.331    & 0.100  & 0.501 & 0.488     & 0.048    & 0.852  \\
                                                                            &                         & Attention-RP            & 0.039                        & 0.235                        & 0.657                     & 0.089    & 0.128  & 0.893 & 0.085    & 0.278  & 0.554 & 0.078     & 0.143    & 0.817  \\
                                                                            &                         & Attention-AT            & 0.082                        & 0.025                        & 0.707                     & 0.006    & 0.157  & 0.907 & 0.035    & 0.230  & 0.510 & 0.049     & 0.193    & 0.818  \\
                                                                            &                         & Attention-iAT           & 0.126                        & 0.228                        & 0.684                     & 0.147    & 0.204  & 0.915 & 0.081    & 0.271  & 0.512 & 0.135     & 0.187    & 0.831  \\
                                                                            &                         & SEAT(\textbf{Ours})     & 3.61E-09                     & 0.001                        & 0.713                     & 2.49E-07 & 0.028  & 0.909 & 8.38E-06 & 0.044  & 0.561 & 1.18E-06  & 0.034    & 0.844  \\ 
\midrule
\multirow{21}{*}{0.05}                                                      & \multirow{7}{*}{RNN}    & Vanilla                 & 0.006                        & 20.911                       & 0.667                     & 0.047    & 28.278 & 0.767 & 0.025    & 18.398 & 0.513 & 0.015     & 21.265   & 0.763  \\
                                                                            &                         & Word-AT                 & 0.023                        & 1.652                        & 0.627                     & 0.019    & 1.270  & 0.797 & 0.031    & 1.320  & 0.522 & 0.030     & 1.126    & 0.748  \\
                                                                            &                         & Word-iAT                & 0.027                        & 1.577                        & 0.637                     & 0.021    & 1.325  & 0.811 & 0.025    & 1.235  & 0.537 & 0.092     & 1.704    & 0.759  \\
                                                                            &                         & Attention-RP            & 0.026                        & 3.294                        & 0.671                     & 0.028    & 2.212  & 0.792 & 0.025    & 2.662  & 0.554 & 0.009     & 3.763    & 0.770  \\
                                                                            &                         & Attention-AT            & 0.055                        & 2.675                        & 0.665                     & 0.048    & 2.611  & 0.782 & 0.032    & 2.210  & 0.528 & 0.068     & 4.370    & 0.755  \\
                                                                            &                         & Attention-iAT           & 0.017                        & 3.559                        & 0.645                     & 0.050    & 2.814  & 0.746 & 0.039    & 2.277  & 0.533 & 0.055     & 1.634    & 0.753  \\
                                                                            &                         & SEAT(\textbf{Ours})     & 1.87E-08                     & 1.546                        & 0.691                     & 3.77E-08 & 1.098  & 0.805 & 4.54E-10 & 1.213  & 0.579 & -3.39E-11 & 3.02E-04 & 0.767  \\ 
\cmidrule{2-15}
                                                                            & \multirow{7}{*}{BiLSTM} & Vanilla                 & 0.003                        & 23.377                       & 0.653                     & 0.258    & 28.472 & 0.809 & 0.190    & 20.458 & 0.524 & 0.022     & 22.146   & 0.764  \\
                                                                            &                         & Word-AT                 & 0.022                        & 1.279                        & 0.656                     & 0.028    & 1.062  & 0.790 & 0.055    & 1.771  & 0.521 & 0.047     & 1.451    & 0.769  \\
                                                                            &                         & Word-iAT                & 0.054                        & 1.900                        & 0.643                     & 0.029    & 1.109  & 0.809 & 0.077    & 1.862  & 0.525 & 0.063     & 1.194    & 0.750  \\
                                                                            &                         & Attention-RP            & 0.031                        & 1.390                        & 0.642                     & 0.036    & 1.409  & 0.772 & 0.054    & 1.381  & 0.522 & 0.067     & 1.533    & 0.764  \\
                                                                            &                         & Attention-AT            & 0.080                        & 1.622                        & 0.672                     & 0.030    & 1.745  & 0.779 & 0.058    & 1.572  & 0.523 & 0.080     & 2.126    & 0.766  \\
                                                                            &                         & Attention-iAT           & 0.035                        & 1.368                        & 0.651                     & 0.036    & 1.712  & 0.801 & 0.063    & 2.314  & 0.525 & 0.078     & 1.907    & 0.777  \\
                                                                            &                         & SEAT(\textbf{Ours})     & 1.44E-08                     & 1.094                        & 0.664                     & 1.04E-08 & 1.043  & 0.801 & 1.47E-08 & 1.229  & 0.524 & 1.35E-08  & 1.027    & 0.768  \\ 
\cmidrule{2-15}
                                                                            & \multirow{7}{*}{BERT}   & Vanilla                 & 0.193                        & 2.197                        & 0.721                     & 0.008    & 2.680  & 0.912 & 0.193    & 1.780  & 0.493 & 0.042     & 2.553    & 0.845  \\
                                                                            &                         & Word-AT                 & 0.004                        & 0.022                        & 0.694                     & 0.175    & 0.065  & 0.910 & 0.111    & 0.058  & 0.546 & 0.473     & 0.056    & 0.836  \\
                                                                            &                         & Word-iAT                & 0.456                        & 0.059                        & 0.658                     & 0.213    & 0.046  & 0.912 & 0.331    & 0.044  & 0.501 & 0.488     & 0.048    & 0.852  \\
                                                                            &                         & Attention-RP            & 0.039                        & 0.235                        & 0.657                     & 0.089    & 0.128  & 0.893 & 0.085    & 0.278  & 0.554 & 0.078     & 0.143    & 0.817  \\
                                                                            &                         & Attention-AT            & 0.082                        & 0.003                        & 0.707                     & 0.006    & 0.157  & 0.907 & 0.035    & 0.230  & 0.510 & 0.049     & 0.193    & 0.818  \\
                                                                            &                         & Attention-iAT           & 0.126                        & 0.228                        & 0.684                     & 0.147    & 0.204  & 0.915 & 0.081    & 0.271  & 0.512 & 0.136     & 0.187    & 0.831  \\
                                                                            &                         & SEAT(\textbf{Ours})     & 1.72E-09                     & 0.001                        & 0.717                     & 1.55E-06 & 0.028  & 0.910 & 8.69E-06 & 0.037  & 0.555 & 1.10E-05  & 0.035    & 0.849  \\
\bottomrule
\end{tabular}
}
\caption{More results on the performance of different methods under different perturbation radii. JSD and TVD are reported to measure the stability while F1 measure the model accuracy of the clean data. From the table we can see that SEAT have lower JSD and TVD with little or even no drop for F1 under all the settings. These suggest that our method suppresses other baseline methods on both model stability and effectiveness. \label{tab: more result main 3 metrics}}
\end{table*}

\begin{table*}[ht]
\small
\centering
\resizebox{1.0\textwidth}{!}{
\begin{tabular}{c|cccccccccccccc} 
\toprule
\multicolumn{1}{c}{\multirow{2}{*}{Radius($\delta_{x}$)}} & \multirow{2}{*}{Model}  & \multirow{2}{*}{Method} & \multicolumn{3}{c}{Emotion}                             & \multicolumn{3}{c}{SST}     & \multicolumn{3}{c}{Hate}    & \multicolumn{3}{c}{RottenT}  \\ 
\cmidrule(l){4-15}
\multicolumn{1}{c}{}                                    &                         &                         & Comp.$\uparrow$ & Suff.$\downarrow$ & Sens.$\downarrow$ & Comp.    & Suff.    & Sens. & Comp.    & Suff.    & Sens. & Comp.    & Suff.    & Sens.  \\ 
\midrule
\multirow{21}{*}{0.005}                                 & \multirow{7}{*}{RNN}    & Vanilla                 & 3.281           & 0.007             & 0.143             & 5.744    & 3.90E-04 & 0.160 & 0.073    & 0.066    & 0.179 & 4.483    & 0.127    & 0.120  \\
                                                        &                         & Word-AT                 & 0.006           & 0.072             & 0.149             & 5.585    & 0.00E+00 & 0.157 & 3.259    & 0.246    & 0.198 & 2.744    & 0.255    & 0.113  \\
                                                        &                         & Word-iAT                & 0.365           & 0.005             & 0.139             & 4.433    & 5.85E-04 & 0.155 & 2.178    & 0.091    & 0.174 & 2.804    & 0.091    & 0.122  \\
                                                        &                         & Attention-RP            & 0.099           & 0.073             & 0.133             & 6.001    & 2.87E-04 & 0.144 & 3.585    & 0.080    & 0.169 & 4.637    & 0.052    & 0.127  \\
                                                        &                         & Attention-AT            & 0.026           & 0.052             & 0.132             & 5.608    & 0.00E+00 & 0.144 & 2.493    & 0.372    & 0.172 & 4.713    & 0.049    & 0.128  \\
                                                        &                         & Attention-iAT           & 1.994           & 6.74E-04          & 0.138             & 4.788    & 0.00E+00 & 0.159 & 5.351    & 0.126    & 0.179 & 2.435    & 0.039    & 0.119  \\
                                                        &                         & SEAT(\textbf{Ours})     & 5.099           & 1.74E-05          & 0.131             & 6.016    & 0.00E+00 & 0.140 & 6.558    & 0.001    & 0.131 & 5.342    & 0.026    & 0.105  \\ 
\cmidrule{2-15}
                                                        & \multirow{7}{*}{BiLSTM} & Vanilla                 & 0.474           & 0.002             & 0.136             & 5.182    & 0.003    & 0.147 & 4.795    & 0.112    & 0.199 & 2.966    & 0.088    & 0.125  \\
                                                        &                         & Word-AT                 & 2.195           & 0.029             & 0.193             & 2.783    & 0.007    & 0.155 & 3.085    & 0.111    & 0.188 & 1.827    & 0.117    & 0.135  \\
                                                        &                         & Word-iAT                & 1.765           & 0.015             & 0.190             & 5.082    & 2.62E-04 & 0.165 & 5.409    & 0.093    & 0.202 & 3.759    & 0.100    & 0.119  \\
                                                        &                         & Attention-RP            & 0.561           & 0.002             & 0.173             & 2.865    & 0.007    & 0.149 & 2.248    & 0.199    & 0.189 & 4.073    & 0.200    & 0.117  \\
                                                        &                         & Attention-AT            & 1.294           & 6.41E-04          & 0.143             & 2.129    & 0.004    & 0.129 & 3.220    & 0.124    & 0.190 & 4.925    & 0.216    & 0.122  \\
                                                        &                         & Attention-iAT           & 0.555           & 0.002             & 0.135             & 5.435    & 0.004    & 0.152 & 4.092    & 0.290    & 0.197 & 4.941    & 0.377    & 0.143  \\
                                                        &                         & SEAT(\textbf{Ours})     & 2.306           & 4.45E-04          & 0.135             & 5.454    & 4.37E-06 & 0.129 & 6.240    & 0.065    & 0.187 & 4.956    & 0.037    & 0.114  \\ 
\cmidrule{2-15}
                                                        & \multirow{7}{*}{BERT}   & Vanilla                 & 5.07E-04        & 0.008             & 0.016             & 0.003    & 0.310    & 0.010 & 3.20E-04 & 0.006    & 0.017 & 0.001    & 0.092    & 0.010  \\
                                                        &                         & Word-AT                 & 6.44E-05        & 0.014             & 0.016             & 0.164    & 0.165    & 0.007 & 4.09E-06 & 0.012    & 0.018 & 4.82E-04 & 0.005    & 0.011  \\
                                                        &                         & Word-iAT                & 4.79E-04        & 0.013             & 0.016             & 9.51E-04 & 0.024    & 0.008 & 3.80E-04 & 0.012    & 0.019 & 6.15E-04 & 0.016    & 0.012  \\
                                                        &                         & Attention-RP            & 0.085           & 0.086             & 0.015             & 0.002    & 0.010    & 0.012 & 0.035    & 0.034    & 0.018 & 0.017    & 0.003    & 0.010  \\
                                                        &                         & Attention-AT            & 4.65E-05        & 0.003             & 0.017             & 5.82E-04 & 0.450    & 0.007 & 0.004    & 0.007    & 0.017 & 0.001    & 0.116    & 0.011  \\
                                                        &                         & Attention-iAT           & 5.74E-05        & 0.164             & 0.016             & 1.19E-04 & 9.07E-04 & 0.007 & 4.05E-04 & 0.151    & 0.017 & 0.003    & 0.025    & 0.011  \\
                                                        &                         & SEAT(\textbf{Ours})     & 0.160           & 0.002             & 0.014             & 0.497    & 0.00E+00 & 0.007 & 0.153    & 6.74E-05 & 0.016 & 0.040    & 0.002    & 0.009  \\ 
\midrule
\multirow{21}{*}{0.01}                                  & \multirow{7}{*}{RNN}    & Vanilla                 & 0.004           & 0.007             & 0.131             & 5.744    & 7.02E-04 & 0.090 & 0.009    & 0.066    & 0.138 & 4.483    & 0.026    & 0.090  \\
                                                        &                         & Word-AT                 & 2.899           & 0.018             & 0.121             & 5.280    & 0.00E+00 & 0.081 & 2.408    & 0.058    & 0.137 & 2.512    & 0.031    & 0.093  \\
                                                        &                         & Word-iAT                & 2.060           & 0.010             & 0.122             & 5.452    & 9.35E-06 & 0.121 & 6.069    & 0.075    & 0.136 & 4.534    & 0.058    & 0.087  \\
                                                        &                         & Attention-RP            & 0.099           & 0.073             & 0.130             & 6.001    & 2.87E-04 & 0.085 & 3.585    & 0.080    & 0.133 & 4.637    & 0.052    & 0.094  \\
                                                        &                         & Attention-AT            & 0.026           & 0.052             & 0.131             & 3.118    & 0.00E+00 & 0.088 & 2.493    & 0.372    & 0.134 & 4.713    & 0.049    & 0.096  \\
                                                        &                         & Attention-iAT           & 1.994           & 6.74E-04          & 0.117             & 4.788    & 0.00E+00 & 0.091 & 5.351    & 0.126    & 0.133 & 2.435    & 0.039    & 0.086  \\
                                                        &                         & SEAT(\textbf{Ours})     & 3.281           & 1.04E-05          & 0.106             & 6.016    & 0.00E+00 & 0.076 & 6.558    & 2.75E-04 & 0.129 & 4.796    & 0.025    & 0.084  \\ 
\cmidrule{2-15}
                                                        & \multirow{7}{*}{BiLSTM} & Vanilla                 & 0.474           & 0.002             & 0.129             & 5.182    & 0.255    & 0.086 & 4.203    & 0.112    & 0.142 & 2.966    & 0.088    & 0.092  \\
                                                        &                         & Word-AT                 & 1.449           & 0.015             & 0.121             & 5.167    & 8.44E-04 & 0.096 & 5.438    & 0.207    & 0.153 & 3.388    & 0.062    & 0.080  \\
                                                        &                         & Word-iAT                & 0.619           & 0.005             & 0.127             & 5.259    & 3.81E-05 & 0.087 & 4.568    & 0.320    & 0.145 & 3.339    & 0.078    & 0.082  \\
                                                        &                         & Attention-RP            & 0.561           & 0.002             & 0.127             & 2.865    & 0.007    & 0.101 & 2.248    & 0.199    & 0.148 & 4.073    & 0.200    & 0.082  \\
                                                        &                         & Attention-AT            & 1.294           & 0.051             & 0.111             & 2.129    & 0.004    & 0.098 & 3.220    & 0.065    & 0.140 & 4.925    & 0.216    & 0.082  \\
                                                        &                         & Attention-iAT           & 0.555           & 0.002             & 0.127             & 5.176    & 0.004    & 0.083 & 4.092    & 0.290    & 0.141 & 2.431    & 0.377    & 0.083  \\
                                                        &                         & SEAT(\textbf{Ours})     & 1.502           & 6.41E-04          & 0.098             & 5.435    & 4.37E-06 & 0.076 & 6.240    & 0.025    & 0.140 & 4.941    & 0.046    & 0.077  \\ 
\cmidrule{2-15}
                                                        & \multirow{7}{*}{BERT}   & Vanilla                 & 5.07E-04        & 0.008             & 0.013             & 0.003    & 0.310    & 0.009 & 3.20E-04 & 0.401    & 0.016 & 0.001    & 0.092    & 0.010  \\
                                                        &                         & Word-AT                 & 5.01E-05        & 0.005             & 0.016             & 1.26E-05 & 4.47E-04 & 0.005 & 4.01E-04 & 0.014    & 0.017 & 0.005    & 0.043    & 0.016  \\
                                                        &                         & Word-iAT                & 5.47E-04        & 0.007             & 0.017             & 1.51E-05 & 5.67E-04 & 0.004 & 0.003    & 0.045    & 0.017 & 2.93E-04 & 0.010    & 0.009  \\
                                                        &                         & Attention-RP            & 0.085           & 0.086             & 0.014             & 0.002    & 0.010    & 0.011 & 0.035    & 0.034    & 0.016 & 0.017    & 0.003    & 0.010  \\
                                                        &                         & Attention-AT            & 4.65E-05        & 0.338             & 0.016             & 4.50E-05 & 0.441    & 0.006 & 0.004    & 0.007    & 0.016 & 6.26E-04 & 0.032    & 0.011  \\
                                                        &                         & Attention-iAT           & 0.002           & 0.164             & 0.015             & 1.19E-04 & 9.07E-04 & 0.007 & 0.001    & 0.151    & 0.017 & 0.003    & 0.025    & 0.010  \\
                                                        &                         & SEAT(\textbf{Ours})     & 0.160           & 0.002             & 0.012             & 0.497    & 0.00E+00 & 0.004 & 0.153    & 0.006    & 0.015 & 0.040    & 0.002    & 0.008  \\ 
\midrule
\multirow{21}{*}{0.05}                                  & \multirow{7}{*}{RNN}    & Vanilla                 & 0.006           & 0.007             & 0.229             & 5.744    & 0.001    & 0.260 & 0.017    & 0.066    & 0.253 & 4.483    & 0.026    & 0.225  \\
                                                        &                         & Word-AT                 & 0.129           & 0.115             & 0.147             & 4.413    & 0.005    & 0.259 & 5.551    & 0.132    & 0.251 & 2.738    & 0.046    & 0.203  \\
                                                        &                         & Word-iAT                & 0.365           & 0.005             & 0.175             & 4.433    & 5.85E-04 & 0.258 & 2.178    & 0.091    & 0.246 & 2.804    & 0.091    & 0.203  \\
                                                        &                         & Attention-RP            & 0.099           & 0.073             & 0.142             & 6.001    & 2.87E-04 & 0.258 & 3.585    & 0.080    & 0.246 & 4.637    & 0.052    & 0.229  \\
                                                        &                         & Attention-AT            & 0.026           & 0.052             & 0.136             & 5.533    & 0.00E+00 & 0.259 & 2.493    & 0.372    & 0.252 & 0.008    & 0.049    & 0.236  \\
                                                        &                         & Attention-iAT           & 1.994           & 6.74E-04          & 0.215             & 4.788    & 0.00E+00 & 0.259 & 5.351    & 0.126    & 0.257 & 2.435    & 0.039    & 0.227  \\
                                                        &                         & SEAT(\textbf{Ours})     & 3.281           & 1.38E-05          & 0.134             & 6.016    & 0.00E+00 & 0.254 & 6.558    & 3.12E-04 & 0.133 & 4.713    & 9.20E-06 & 0.132  \\ 
\cmidrule{2-15}
                                                        & \multirow{7}{*}{BiLSTM} & Vanilla                 & 0.474           & 0.002             & 0.258             & 5.182    & 1.71E-04 & 0.258 & 4.916    & 0.112    & 0.260 & 2.966    & 0.088    & 0.224  \\
                                                        &                         & Word-AT                 & 1.756           & 0.006             & 0.259             & 4.939    & 0.009    & 0.260 & 2.743    & 0.115    & 0.259 & 2.666    & 0.137    & 0.228  \\
                                                        &                         & Word-iAT                & 1.765           & 0.025             & 0.259             & 5.082    & 2.62E-04 & 0.261 & 5.409    & 0.093    & 0.259 & 3.759    & 0.100    & 0.228  \\
                                                        &                         & Attention-RP            & 0.561           & 0.002             & 0.179             & 2.865    & 0.007    & 0.250 & 2.248    & 0.199    & 0.258 & 4.073    & 0.200    & 0.225  \\
                                                        &                         & Attention-AT            & 1.294           & 6.41E-04          & 0.252             & 2.129    & 0.004    & 0.259 & 3.220    & 0.178    & 0.259 & 4.925    & 0.216    & 0.236  \\
                                                        &                         & Attention-iAT           & 0.555           & 0.002             & 0.178             & 4.869    & 0.004    & 0.258 & 4.092    & 0.290    & 0.260 & 3.364    & 0.377    & 0.250  \\
                                                        &                         & SEAT(\textbf{Ours})     & 1.849           & 4.45E-04          & 0.169             & 5.435    & 4.37E-06 & 0.235 & 6.240    & 0.065    & 0.257 & 4.941    & 0.062    & 0.212  \\ 
\cmidrule{2-15}
                                                        & \multirow{7}{*}{BERT}   & Vanilla                 & 5.07E-04        & 0.008             & 0.021             & 0.003    & 0.310    & 0.015 & 3.20E-04 & 0.141    & 0.029 & 0.001    & 0.092    & 0.021  \\
                                                        &                         & Word-AT                 & 2.55E-05        & 0.012             & 0.017             & 7.64E-04 & 0.020    & 0.021 & 7.21E-04 & 0.023    & 0.026 & 7.75E-04 & 0.009    & 0.020  \\
                                                        &                         & Word-iAT                & 4.79E-04        & 0.013             & 0.024             & 9.51E-04 & 0.024    & 0.015 & 3.80E-04 & 0.012    & 0.028 & 6.15E-04 & 0.016    & 0.022  \\
                                                        &                         & Attention-RP            & 0.085           & 0.086             & 0.021             & 0.002    & 0.010    & 0.018 & 0.035    & 0.034    & 0.022 & 0.017    & 0.003    & 0.016  \\
                                                        &                         & Attention-AT            & 4.65E-05        & 0.003             & 0.017             & 7.31E-04 & 0.467    & 0.016 & 0.004    & 0.007    & 0.023 & 0.003    & 0.304    & 0.018  \\
                                                        &                         & Attention-iAT           & 6.23E-05        & 0.164             & 0.020             & 1.19E-04 & 9.07E-04 & 0.016 & 3.18E-04 & 0.151    & 0.021 & 0.003    & 0.025    & 0.017  \\
                                                        &                         & SEAT(\textbf{Ours})     & 0.160           & 0.002             & 0.017             & 0.497    & 0.00E+00 & 0.012 & 0.153    & 0.006    & 0.019 & 0.040    & 0.002    & 0.015  \\
\bottomrule
\end{tabular}
}
\caption{More results on the performance of different methods under different perturbation radii. Comprehensiveness, Sufficiency, and Sensitivity are reported to measure the model interpretability. From the table we can see that SEAT outperforms baseline models under all three metrics. It suggest that our method outperforms other baseline methods on model interpretability. \label{tab: more result other three metrics}}
\end{table*} 

\subsection{More Results on Stability of Attention under Word Perturbation}
In this section, we show results on the stability of different methods under word perturbation. The perturbation word number $N$ are set as 2 and 3. As we can see from Tab.\ref{tab_all_wordp}, across all the setting, our method achieves the best performance compared to all baseline methods. 

\begin{table*}[htbp]
\centering
\resizebox{\textwidth}{!}{
\begin{tabular}{cccccccccccccc} 
\toprule
\multirow{2}{*}{Perturb Word $N$} & \multirow{2}{*}{Method} & \multicolumn{3}{c}{Emotion}                                                             & \multicolumn{3}{c}{SST}  & \multicolumn{3}{c}{Hate} & \multicolumn{3}{c}{RottenT}  \\ 
\cmidrule(lr){3-5}\cmidrule(r){6-8}\cmidrule(lr){9-11}\cmidrule(lr){12-14}
                                  &                         & JSD$\downarrow$ & TVD$\downarrow$ & F1$\uparrow$ & JSD      & TVD   & F1    & JSD      & TVD   & F1    & JSD      & TVD   & F1        \\ 
\midrule
\multirow{7}{*}{$N$=2}              & Vanilla                 & 0.628                        & 2.861                        & 0.721                     & 0.343    & 3.806 & 0.912 & 0.435    & 2.098 & 0.493 & 0.582    & 3.570 & 0.845     \\
                                  & Word-AT                 & 0.004                        & 0.022                        & 0.694                     & 0.175    & 0.065 & 0.910 & 0.111    & 0.058 & 0.546 & 0.473    & 0.056 & 0.836     \\
                                  & Word-iAT                & 0.456                        & 0.059                        & 0.658                     & 0.213    & 0.046 & 0.912 & 0.331    & 0.044 & 0.501 & 0.488    & 0.048 & 0.852     \\
                                  & Attention-RP            & 0.039                        & 0.235                        & 0.657                     & 0.089    & 0.128 & 0.893 & 0.085    & 0.278 & 0.554 & 0.078    & 0.143 & 0.817     \\
                                  & Attention-AT            & 0.082                        & 0.003                        & 0.707                     & 0.006    & 0.157 & 0.907 & 0.035    & 0.230 & 0.510 & 0.049    & 0.193 & 0.818     \\
                                  & Attention-iAT           & 0.126                        & 0.228                        & 0.684                     & 0.147    & 0.204 & 0.915 & 0.081    & 0.271 & 0.512 & 0.136    & 0.187 & 0.831     \\
                                 \rowcolor{grey!20} & SEAT (\textbf{Ours})    & 1.72E-09                     & 0.001                        & 0.716                     & 1.55E-06 & 0.028 & 0.906 & 8.69E-06 & 0.037 & 0.555 & 1.10E-05 & 0.035 & 0.849     \\ 
\midrule
\multirow{7}{*}{$N$=3}              & Vanilla                 & 0.618                        & 2.864                        & 0.721                     & 0.326    & 3.852 & 0.912 & 0.387    & 2.156 & 0.493 & 0.564    & 3.610 & 0.845     \\
                                  & Word-AT                 & 0.004                        & 0.022                        & 0.694                     & 0.175    & 0.065 & 0.910 & 0.111    & 0.058 & 0.546 & 0.473    & 0.056 & 0.836     \\
                                  & Word-iAT                & 0.456                        & 0.059                        & 0.658                     & 0.213    & 0.046 & 0.912 & 0.331    & 0.044 & 0.501 & 0.488    & 0.048 & 0.852     \\
                                  & Attention-RP            & 0.039                        & 0.235                        & 0.657                     & 0.089    & 0.128 & 0.893 & 0.085    & 0.278 & 0.554 & 0.078    & 0.143 & 0.817     \\
                                  & Attention-AT            & 0.082                        & 0.003                        & 0.707                     & 0.006    & 0.157 & 0.907 & 0.035    & 0.230 & 0.510 & 0.049    & 0.193 & 0.818     \\
                                  & Attention-iAT           & 0.126                        & 0.228                        & 0.684                     & 0.147    & 0.204 & 0.915 & 0.081    & 0.271 & 0.512 & 0.136    & 0.187 & 0.831     \\ 
                                 \rowcolor{grey!20} &  SEAT (\textbf{Ours})    & 1.72E-09                     & 0.001                        & 0.715                     & 1.55E-06 & 0.028 & 0.909 & 8.69E-06 & 0.037 & 0.555 & 1.10E-05 & 0.035 & 0.846     \\
\bottomrule
\end{tabular}}
\caption{More Results of evaluating stability of different methods under word perturbation.}
\label{tab_all_wordp}
\end{table*}

\section{More Examples under Different Perturbation}
In this section, we show more real examples to illustrate the stability of the proposed method under different kinds of perturbations. BERT is used as the model architecture and IMDB \cite{maas-EtAl:2011:ACL-HLT2011} is used as the demonstrated dataset. 

\begin{figure*}[!thbp]
\centering
\includegraphics[width=.9\columnwidth]{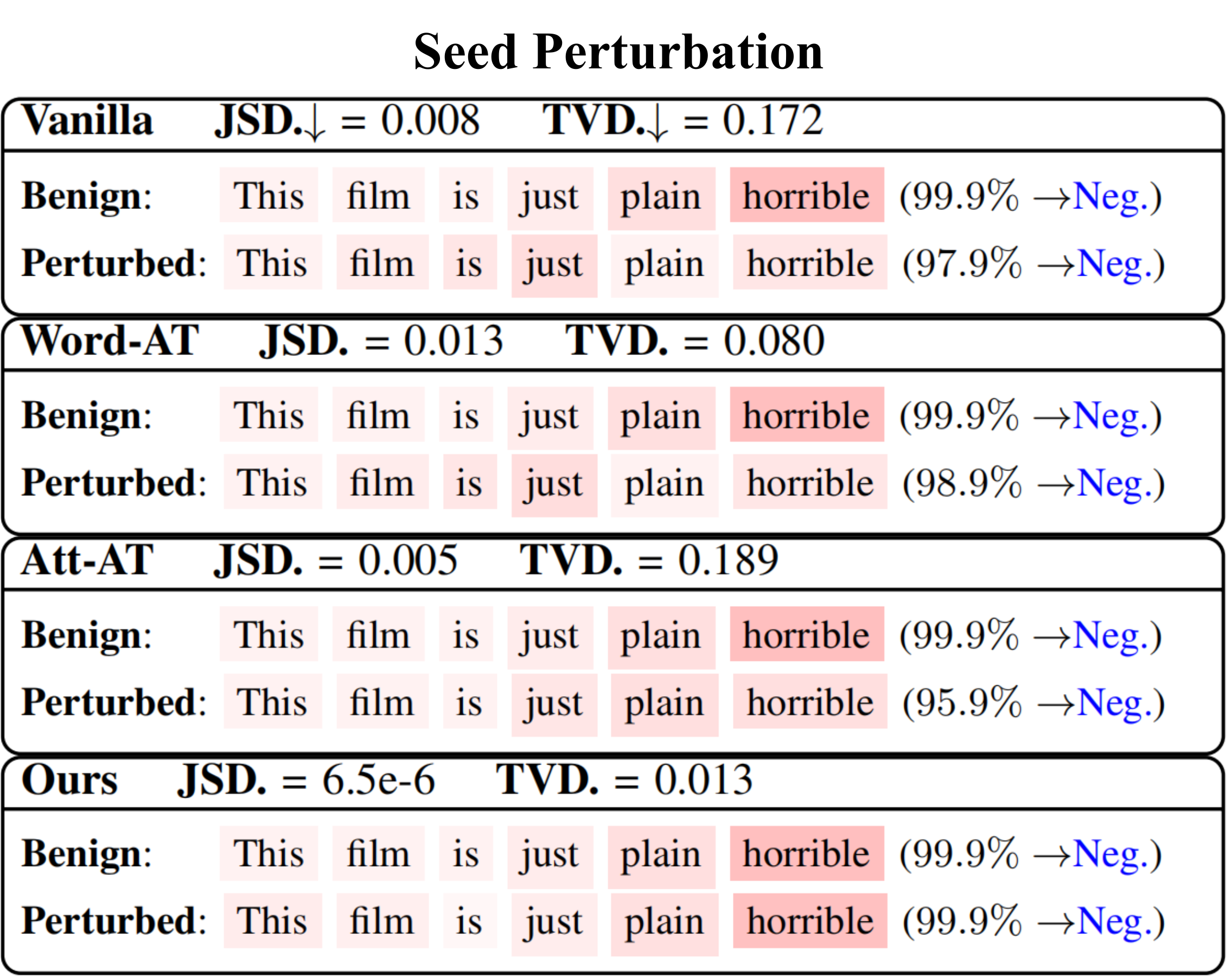}
\caption{
An example of the stability of model trained with different method under seed perturbation. As we can see from figure, the perturbation of random seed mainly affect the attention distribution while producing little change on the prediction. Compared to baseline methods, our method generate nearly similar attention weight distribution as it is before perturbed. This observation indicates that our propose method are able to improve the stability under seed perturbation.
}
\label{fig:exampleseed}
\end{figure*}

As we can see from Fig.\ref{fig:exampleseed},  different random seeds mainly affect the attention weight distribution (or explainable heat map in Fig.\ref{fig:exampleseed})  while produce little change on the prediction. Compared to baseline methods, the attention weight distribution generated by our method are more stable under perturbation.

\begin{figure*}[!thbp]
\centering
\includegraphics[width=.9\columnwidth]{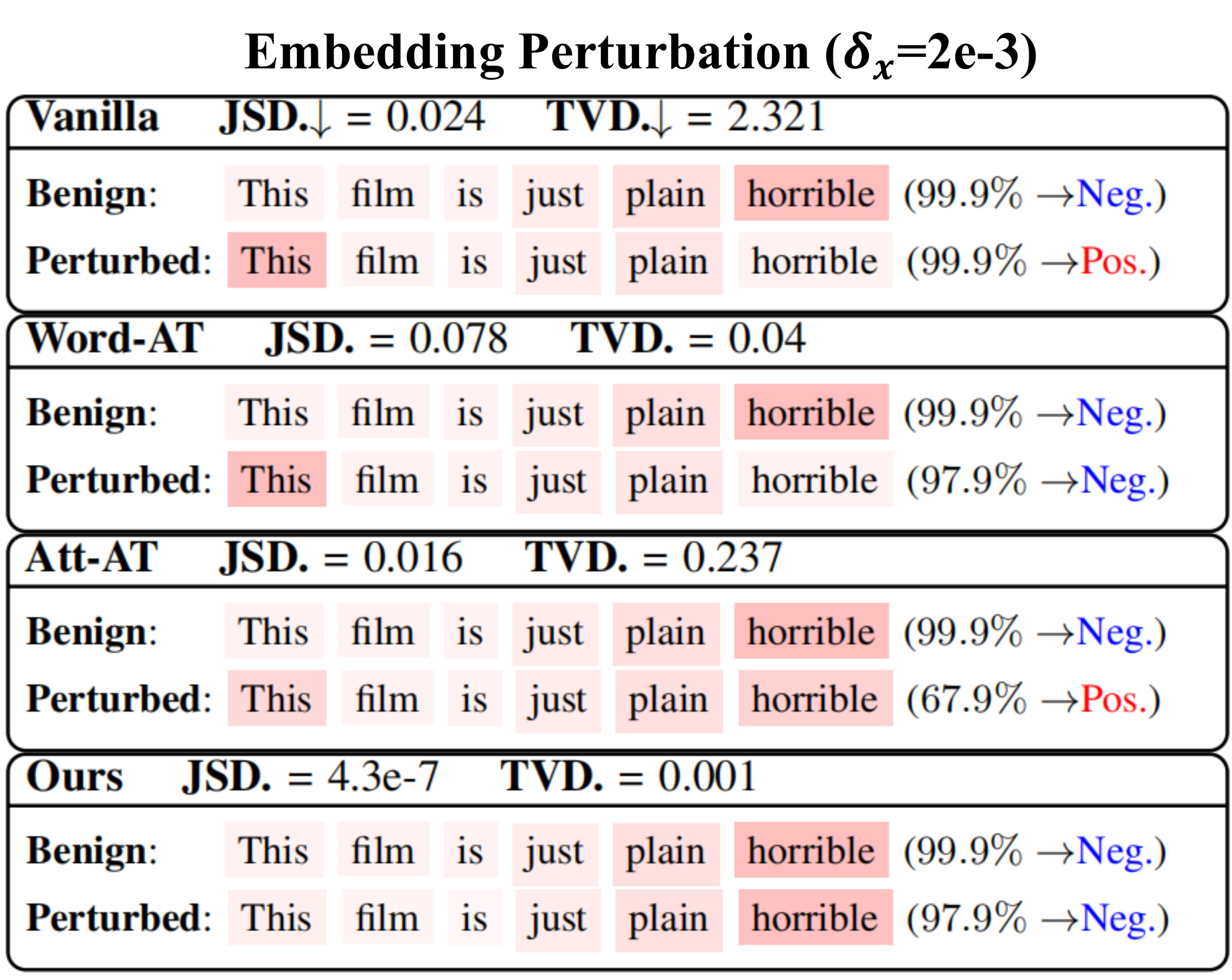}
\caption{
An example of the stability of model trained with different method under embedding perturbation. As it is shown in the figure, we can observe more obvious change both on model predication and attention weight distribution. The considered baseline methods are either not sufficiently stable in terms of prediction or not stable in terms of attention distribution. Results demonstrate that our method outperform them in terms of stability under embedding perturbation.
}
\label{fig:exampleembedding}
\end{figure*}

For the results under perturbation on embedding space, we can observe more notable change  on both  model predication and attention weight distribution. As we can see from Fig.\ref{fig:exampleembedding}, vanilla attention methods are unstable under perturbation even when $\delta_x=$2e-3 since it classifies the input text to wrong class with high probability. Despite that model trained with Word-AT can ensure the stability on prediction, however, it still produces unstable attention weight distribution. Contrast to Word-AT, Attention-AT can  improve the stability of attention distribution while lack stability on prediction. Compared to these baseline methods, our methods achieve the best performance under two metrics. It can be seen from Fig.\ref{fig:exampleembedding} that model trained with our method is stable under perturbation on both the prediction and the attention distribution. 

\section{Additional Ablation Study}
In this section, we provide more ablation studies on the exaplianability and stability of model. We evaluate each module (regularization) in Eq.\ref{eq:6}. As we can see from Tab.\ref{tab:ablation-2}, $\mathcal{L}_3$ are more effective in reducing Sens. while $\mathcal{L}_{Topk}$ contribute more on the Comp. and Suff. And the combination of all the competent yield the model with the highest exaplianability. This indicates that the loss term $\mathcal{L}_{Topk}$ and $\mathcal{L}_3$ proposed in Eq.\ref{eq:6} are both indispensable for the performance improvement on explainability.

Tab. \ref{tab:ablation} further confirms that each regularizer in our objective function is indispensable. Specifically, we can see JSD will decrease significantly if we add the $\mathcal{L}_{Topk}$ loss. This is due to that 
$\mathcal{L}_{Topk}$ enforces  a large  overlaps on top-k indices and thus it makes SEAT inherit the  explainability of vanilla attention. Moreover, in the case where there is $\mathcal{L}_{Topk}$, adding  term $\mathcal{L}_3$ could further decrease JSD as it further enhances the stability.

\end{document}